\documentclass[lettersize,journal]{IEEEtran}
\usepackage{amsmath,amsfonts}
\usepackage{amssymb}
\usepackage{algorithmic}
\usepackage{algorithm}
\usepackage{array}
\usepackage[caption=false,font=normalsize,labelfont=sf,textfont=sf]{subfig}
\usepackage{textcomp}
\usepackage{stfloats}
\usepackage{url}
\usepackage{verbatim}
\usepackage{graphicx}
\usepackage{cite}
\usepackage{threeparttable}
\usepackage{arydshln}
\usepackage{cuted}
\usepackage{caption}
\usepackage{multirow}
\usepackage{ragged2e}
\usepackage{tabularx}
\usepackage{pifont}
\usepackage{makecell}
\usepackage{xcolor}
\usepackage{float}
\newcommand{\blue}[1]{\textcolor{black}{#1}}

\begin{document}

\title{\blue{DreamBarbie}: Text to Barbie-Style 3D Avatars}

\author{Xiaokun Sun, Zhenyu Zhang$^{*}$, Ying Tai, Hao Tang, Zili Yi, Jian Yang

\thanks{$^{*}$Corresponding Author: Zhenyu Zhang (Email: zhangjesse@foxmail.com).}
\thanks{Xiaokun Sun, Zhenyu Zhang, Ying Tai, Zili Yi, and Jian Yang are affiliated with the School of Intelligence Science and Technology, Nanjing University, Jiangsu 215163, China.~Email: xiaokun\_sun@smail.nju.edu.cn, zhangjesse@foxmail.com, \{yingtai, yi, csjyang\}@nju.edu.cn.}
\thanks{Hao Tang is affiliated with the School of Computer Science, Peking University, Beijing 100871, China. Email: bjdxtanghao@gmail.com.}
\thanks{This work was supported by the NSFC under Grant No. 62376121, Basic Research Program of Jiangsu under Grant No. BK20251999, Gusu Innovation Leading Talent Program under Grant No. ZXL2025319, and Jiangsu Provincial Science \& Technology Major Project under Grant No. BG2024042.}

}

\markboth{Journal of \LaTeX\ Class Files,~Vol.~14, No.~8, August~2021}%
{Sun \MakeLowercase{\textit{et al.}}: \blue{DreamBarbie}: Text to Barbie-Style 3D Avatars}


\maketitle
\vspace{-2cm}
\begin{strip}
    \centering
    \vspace*{-4cm}
    \includegraphics[width=\textwidth]{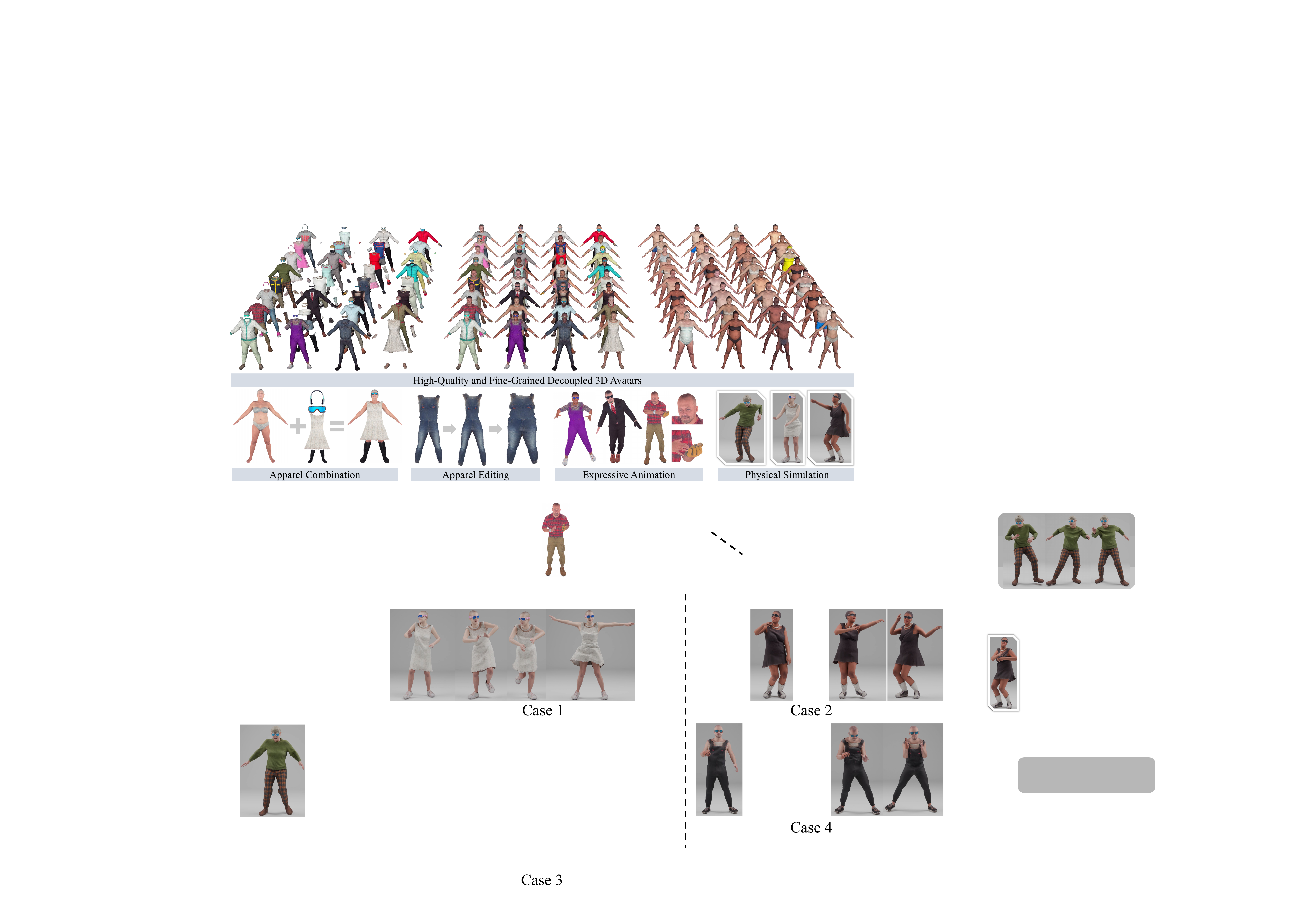} 
    \captionof{figure}{Our method generates Barbie-style 3D avatars from textual input. ``\textbf{Barbie-style}'' refers to the following key characteristics: \textbf{(1) High-Quality} geometry and realistic appearance, ensuring visually lifelike avatars; \textbf{(2) Fine-Grained Decoupling}, separating body, clothing, shoes, and accessories to enable flexible apparel combination and editing; \textbf{(3) Expressive Animation}, supporting a wide range of body movements, facial expressions, and hand gestures; \textbf{(4) Simulation Compatibility}, enabling modeling of non-watertight garments and seamless integration into existing physical simulation pipelines.}
    \label{fig:teaser}
    \vspace{-0.5cm}
\end{strip}

\begin{abstract}
To integrate digital humans into everyday life, there is a strong demand for generating high-quality, fine-grained disentangled 3D avatars that support expressive animation and simulation capabilities, ideally from low-cost textual inputs.
Although text-driven 3D avatar generation has made significant progress by leveraging 2D generative priors, existing methods still struggle to fulfill all these requirements simultaneously.
To address this challenge, we propose \blue{DreamBarbie}, a novel text-driven framework for generating animatable 3D avatars with separable shoes, accessories, and simulation-ready garments, truly capturing the iconic ``Barbie doll'' aesthetic.
The core of our framework lies in an expressive 3D representation combined with appropriate modeling constraints.
\blue{Unlike prior methods, we use G-Shell to uniformly model watertight components (e.g., bodies, shoes) and non-watertight garments.
By reformulating boundaries as Euclidean field intersections instead of manifold geodesics, we propose an SDF-based initialization and a hole regularization loss that together achieve a $100\times$ speedup and stable open topology without image input.}
These disentangled 3D representations are then optimized by specialized expert diffusion models tailored to each domain, ensuring high-fidelity outputs.
To mitigate geometric artifacts and texture conflicts when combining different expert models, we further propose several effective geometric losses and strategies.
Extensive experiments demonstrate that \blue{DreamBarbie} outperforms existing methods in both dressed human and outfit generation.
Our framework further enables diverse applications, including apparel combination, editing, expressive animation, and physical simulation.
Project page: \url{https://xiaokunsun.github.io/DreamBarbie.github.io/}.
\end{abstract}


\vspace{-0.5cm}
\section{Introduction}
\label{sec:1}
Automating the creation of 3D avatars without manual effort remains a critical challenge in computer vision and graphics, with applications ranging from VR/AR and virtual try-ons to gaming and beyond~\cite{vr1,vr2,vr3}.
To fully unlock the potential of these applications, it is essential that the generated avatars meet several key standards: \textbf{(1) High Quality}: Digital avatars should feature exquisite geometry and realistic appearances; \textbf{(2) Fine-Grained Decoupling}: Enabling separation of body, clothing, shoes, and accessories within the avatar model for flexible customization and editing; \textbf{(3) Expressive Animation}: Supporting animations through body movements, facial expressions, and hand gestures; \textbf{(4) Simulation Compatibility}: Ensuring that digital avatars dress in non-watertight garments, integrating seamlessly with existing physical simulation pipelines.
In vivid terms, we expect the generated digital human with qualities reminiscent of \textbf{Barbie dolls}, thereby enhancing their practical value across application scenarios.

Recently, text-driven 3D avatar generation methods~\cite{avatarclip,dreamhuman,humannorm}, which integrate pretrained text-to-image (T2I) diffusion models~\cite{t2i1,latentdiffusion}, have gained significant attention by eliminating the need for human image/video/3D data collection~\cite{pifu,octopus,innerbody,coma,ganhead,gdna}.
However, as summarized in Table~\ref{tab:summary}, existing text-to-avatar methods fail to meet the above ``Barbie-like'' standard.
Specifically, approaches~\cite{avatarclip,dreamhuman,dreamwaltz,avatarverse,dreamavatar,avatarfusion,humancoser,tela} leveraging implicit NeRF~\cite{nerf,neus} cannot support physical simulation or accurately model facial expressions and hand movements due to a lack of explicit structures.
Moreover, these methods tend to produce overly smooth surfaces.
Methods~\cite{humangaussian,gavatar,dreamwaltzg,laga,simavatar,tada,x-oscar,star,so-smpl} using explicit 3DGS~\cite{3dgs} and SMPL-X~\cite{smpl-x} enhance animation expressiveness and simulation compatibility but struggle to capture delicate geometric details.
Approaches~\cite{humannorm,seeavatar} based on the hybrid representation DMTet~\cite{dmtet}, integrating strengths of both implicit and explicit 3D representations (\textit{i.e.}, stable optimization properties and explicit structure), are capable of modeling detailed geometry. 
However, DMTet cannot generate non-watertight surfaces required for realistic garment simulation.
Furthermore, current text-to-disentangled-avatar methods typically adopt a single general diffusion model to guide both body and apparel generation.
This general constraint compromises in-domain fidelity and fails to produce diverse outfits (\textit{e.g.}, shoes, necklaces, glasses, or other accessories).

It is natural to ask: \textit{Can we automatically generate Barbie-like 3D avatars based on text?} 
Our answer is yes, but this is a non-trivial task due to several challenges.
\textbf{(1)} How can we model both watertight components (\textit{e.g.}, bodies, shoes, and accessories) and non-watertight, simulation-ready garments within a unified framework?
While more expressive representations (\textit{e.g.}, G-Shell~\cite{gshell}) offer potential solutions, initializing and regularizing them to accurately represent open surfaces \textbf{without multi-view image inputs} remains an open problem.
\textbf{(2)} Providing suitable constraints for optimizing decoupled bodies and outfits to achieve domain-specific realism is both key and challenging.

\begin{table}[tp]
    \renewcommand{\arraystretch}{1.2}
    \begin{center}
    \caption{Comparisons of Existing Text-Driven 3D Avatar Generation Methods}
    \label{tab:summary}
    \resizebox{0.48\textwidth}{!}{
    \begin{threeparttable}
    \begin{tabular}{c|c|c|c|c|c}
    \hline
    \textbf{Methods} & \textbf{Representation} & \makecell{\textbf{High}\\\textbf{Quality}} & \makecell{\textbf{Fine-Grained}\\\textbf{Decoupling}} & \makecell{\textbf{Expressive}\\\textbf{Animation}} & \makecell{\textbf{Simulation}\\\textbf{Compatibility}} \\
    \hline  
    \cite{avatarclip,dreamhuman,dreamwaltz,avatarverse,dreamavatar} & NeRF/Neus & \ding{56} & \ding{56} & \ding{56} & \ding{56} \\
    \cite{avatarfusion,humancoser,tela} & NeRF/Neus & \ding{56} & \ding{52}\rotatebox[origin=c]{-9.2}{\kern-0.7em\ding{55}} & \ding{56} & \ding{56} \\
    \cdashline{1-6}
    \cite{humangaussian,gavatar} & 3DGS & \ding{52}\rotatebox[origin=c]{-9.2}{\kern-0.7em\ding{55}} & \ding{56} & \ding{56} & \ding{56} \\
    \cite{dreamwaltzg} & 3DGS & \ding{52}\rotatebox[origin=c]{-9.2}{\kern-0.7em\ding{55}} & \ding{56} & \ding{52} & \ding{56} \\
    \cite{laga} & 3DGS & \ding{56} & \ding{52}\rotatebox[origin=c]{-9.2}{\kern-0.7em\ding{55}} & \ding{56} & \ding{56} \\
    \cite{simavatar}$^{\dagger}$ & 3DGS & \ding{56} & \ding{52} & \ding{56} & \ding{52} \\
    \cdashline{1-6}
    \cite{tada,x-oscar,star} & SMPL-X & \ding{52}\rotatebox[origin=c]{-9.2}{\kern-0.7em\ding{55}} & \ding{56} & \ding{52} & \ding{56} \\
    \cite{so-smpl} & SMPL-X & \ding{56} & \ding{52}\rotatebox[origin=c]{-9.2}{\kern-0.7em\ding{55}} & \ding{52} & \ding{52} \\
    \cdashline{1-6}
    \cite{humannorm,seeavatar,avatarstudio} & DMTet & \ding{52} & \ding{56} & \ding{56} & \ding{56} \\
    \cdashline{1-6}
    \blue{DreamBarbie} (Ours) & G-Shell & \ding{52} & \ding{52} & \ding{52} & \ding{52} \\
    \hline
    \end{tabular}
    \begin{tablenotes}
        \item $\dagger$ Although SimAvatar~\cite{simavatar} does not separate the body, clothing, shoes, and accessories, it disentangles the body, clothing, and hair. Therefore, we consider it to have achieved fine-grained decoupling.
    \end{tablenotes}
    \end{threeparttable}
    }
    \end{center}
    \vspace{-0.8cm}
\end{table}

To address these challenges, we introduce \textbf{\blue{DreamBarbie}}\footnote{\blue{Barbie\textsuperscript{\textregistered} is a registered trademark of Mattel, Inc. This project is for academic research purposes only and is not affiliated with, endorsed by, or connected to Mattel, Inc.}}, a novel text-driven framework for generating Barbie-style 3D avatars. 
As shown in Fig.~\ref{fig:teaser}, avatars from \blue{DreamBarbie} not only exhibit exquisite geometry and textures but also feature a variety of decoupled shoes, accessories, and simulation-ready garments that can be freely combined and edited.
Furthermore, our framework supports expressive animation and compatibility with physical simulations, thereby satisfying all four aforementioned requirements simultaneously.

The core of the framework consists of two key components:
\blue{\textbf{(1)} Unlike prior methods, we utilize expressive G-Shell~\cite{gshell} to unify watertight and non-watertight modeling.
We propose an SDF-based boundary initialization that avoids manifold geodesics via a 3D proxy (Pie Mesh), ensuring stable gradients and $100\times$ acceleration. 
Combined with a hole-preserving loss, this enables clean open surface modeling without multi-view supervision.}
\textbf{(2)} Instead of relying on a single general model as in prior text-to-disentangled-avatar methods, we appropriately incorporate different expert diffusion models to guarantee the domain-specific fidelity.
A series of effective regularization losses and strategies is also proposed to address geometric artifacts and texture conflicts when combining different expert models.
Specifically, our method generates Barbie-like 3D digital humans in three stages: 
First, we generate a reasonable and realistic base body using human-specific diffusion models along with a proposed SMPLX-evolving prior loss.
Second, we initialize apparel with the semantic-aligned body and optimize it using object-specific generative priors and several solid geometric losses.
Finally, we jointly fine-tune the assembled avatar to enhance texture harmony and consistency.
Through this pipeline, the generated 3D avatars are animatable and disentangled, featuring high-quality bodies, shoes, accessories, and simulation-ready garments, delivering a truly immersive ``\textbf{Barbie-style}'' digital experience.

Our main contributions are summarized as follows: 
\begin{itemize}
    \item We introduce \blue{DreamBarbie}, a novel text-driven framework for generating realistic and highly disentangled 3D avatars. The framework enables the decoupling of bodies, garments, shoes, and accessories, while enabling outfit transfer, editing, expressive animation, and simulation.
    \item We innovatively adopt G-Shell to uniformly model watertight and non-watertight components with high fidelity. We further propose an efficient initialization strategy and a hole-preserving loss to ensure clean open surface modeling. To the best of our knowledge, this is the first work to incorporate G-Shell into text-to-3D generation.
    \item We effectively integrate expert models at different optimization stages to provide suitable guidance, significantly improving the in-domain realism of generated components. Additionally, we introduce a series of geometric losses and strategies to address geometric artifacts and texture conflicts when combining different expert models.
    \item Extensive experiments demonstrate that \blue{DreamBarbie} outperforms existing methods in avatar and outfit generation, achieving superior results in geometry quality, texture detail, and alignment with text descriptions.
\end{itemize}

\vspace{-0.5cm}
\section{Related Work}
\label{sec:2}
\noindent
\textit{Text to Holistic 3D Avatar Generation.}
The field of text-to-3D generation has witnessed significant progress~\cite{dreamfusion,fantasia3d,gaussiandreamer,dreamgaussian,prolificdreamer,luciddreamer,richdreamer,mvdream}, largely driven by advances in text-to-image models~\cite{t2i1,latentdiffusion,controlnet}.
Building upon the success in generating general 3D objects, various methods~\cite{avatarclip,dreamhuman,dreamwaltz,avatarverse,dreamavatar,humangaussian,gavatar,dreamwaltzg,tada,x-oscar,star,humannorm,seeavatar} have been proposed to generate complex 3D avatars.
AvatarCLIP~\cite{avatarclip} pioneers zero-shot generation of 3D digital humans from text prompts by leveraging human priors and CLIP~\cite{clip}.
With the introduction of the Score Distillation Sampling (SDS) loss~\cite{dreamfusion}, several subsequent works~\cite{dreamhuman,dreamwaltz,avatarverse,dreamavatar} combine it with parametric human models~\cite{smpl,smpl-x,imghum} to significantly improve the fidelity of generated avatars.
To enhance animation expressiveness and simulation compatibility, approaches~\cite{humangaussian,gavatar,dreamwaltzg,tada,x-oscar,star} replace implicit NeRF~\cite{nerf,neus} with explicit representations such as 3DGS~\cite{3dgs} and SMPL-X~\cite{smpl-x} for modeling 3D humans.
However, purely implicit or explicit representations struggle to capture complex geometric details, limiting the quality of generated digital humans.
Accordingly, HumanNorm~\cite{humannorm}, SeeAvatar~\cite{seeavatar}, and AvatarStudio~\cite{avatarstudio} propose using the hybrid representation DMTet~\cite{dmtet}, which combines the advantages of both implicit and explicit representations to achieve high-fidelity avatar generation.
Nevertheless, these methods typically model the human body, garments, shoes, and accessories as a single model, sacrificing flexibility and controllability for downstream applications such as apparel composition and physical simulation.

\textit{Text to Disentangled 3D Avatar Generation.}
To enable controllable avatar creation, several disentangled approaches have been proposed that separately model the human body and outfits through multi-stage optimization.
Humancoser~\cite{humancoser} and AvatarFusion~\cite{avatarfusion} use two implicit NeRF~\cite{nerf,neus} to represent the body and clothing, respectively.
LAGA~\cite{laga} and TELA~\cite{tela} adopt a layer-wise framework, modeling each clothing item as an independent layer, further improving generation flexibility and controllability.
SO-SMPL~\cite{so-smpl} and SimAvatar~\cite{simavatar} leverage sequentially offset SMPL-X~\cite{smpl-x} and compositional 3DGS~\cite{3dgs} anchored on the mesh to enhance compatibility with simulation pipelines.
However, these methods rely on a single general T2I model to guide both body and clothing generation, leading to compromised fidelity in geometry or texture within specific domains.
Moreover, they struggle to generate diverse accessories such as shoes, necklaces, glasses, or other fine-grained items.

In contrast, avatars from \blue{DreamBarbie} not only exhibit exquisite geometry and appearance but also wear multiple separable, realistic outfits, while supporting expressive animation and physical simulation.
We summarize the main differences between \blue{DreamBarbie} and existing methods in Table~\ref{tab:summary}.

\textit{3D Representation of Decoupled Avatar Modeling.}
The disentangled modeling of the human body and clothing has been extensively studied in computer vision and graphics. 
Depending on specific problem settings, different representations are chosen based on trade-offs among optimization properties, geometric modeling capabilities, animation expressiveness, and simulation compatibility:
\textbf{(1) Implicit representations}~\cite{nerf,neus,sdf,udf} are widely adopted in decoupled 3D avatar reconstruction and generation tasks~\cite{humancoser,avatarfusion,tela,neuralabc} due to their stable optimization properties. However, they lack explicit structures, making them unsuitable for expressive animation and physical simulation.
\textbf{(2) Explicit representations}~\cite{3dgs,smpl,smpl-x} are commonly used in disentangled digital human creation~\cite{laga,simavatar,so-smpl,bcnet,layga}, as they provide explicit structures suitable for animation and simulation. Nevertheless, they struggle to capture delicate geometric details.
\textbf{(3) Combinations of the above representations} have also been proposed to achieve high-fidelity reconstruction and generation of decoupled 3D avatars~\cite{scarf,teca,delta,smplicit}. 
Despite these efforts, the inherent disadvantages mentioned above persist.
\textbf{(4) Hybrid representations}~\cite{dmtet,gshell} combine the advantages of implicit and explicit representations to sculpt impressive geometric details in reconstructed disentangled 3D avatars~\cite{gala,d3human}.
G-Shell~\cite{gshell}, in particular, addresses the limitation that DMTet~\cite{dmtet} cannot model non-watertight surfaces compatible with physical simulations.
However, how to initialize and regularize G-Shell without strong multi-view image input remains an open problem.
In this work, we aim to unlock the potential of G-Shell for the challenging task of text-driven Barbie-style 3D avatar generation.

\vspace{-0.2cm}
\section{Method}
\label{sec:3}

\subsection{Preliminary}
\label{sec:3.1}
\noindent
\textit{SMPL-X}~\cite{smpl-x} is a parametric human model that represents the shape, pose, and expression using a set of parameters.
Given shape parameters $\beta$, body pose parameters ${\theta}_{body}$, jaw pose parameters ${\theta}_{jaw}$, hand pose parameters ${\theta}_{hand}$, and expression parameters $\psi_{exp}$, it generates a 3D human mesh ${M}_{smplx}$.

\textit{Score Distillation Sampling}~\cite{dreamfusion} leverages a pre-trained T2I model to guide the alignment of a 3D representation with input text.
Given a text prompt $y$, a 3D representation parameterized by $\theta$, and a diffusion model parameterized by $\phi$, the SDS loss is defined as:
\begin{equation}
    \nabla_{\theta} \mathcal{L}_{SDS} = \mathbb{E}_{t, \epsilon} \left[ w(t) ({\epsilon}_{\phi}(x_t; y, t) - \epsilon ) \frac{\partial x}{\partial \theta} \right],
    \label{eq:sds}
\end{equation}
where $t$ is the time step in the 2D diffusion process, $x = g(\theta)$ is the image rendered from $\theta$ by a differentiable renderer~\cite{renderer} $g(\cdot)$, $x_t = x + \epsilon$ is a noised version of $x$, ${\epsilon}_{\phi}(x_t; y, t)$ is the denoised image, and $w(t)$ is the weight function. For simplicity, we omit $w(t)$ in the following formulas.

\textit{DMTet}~\cite{dmtet} is a hybrid representation that combines an implicit signed distance field (SDF) $s: \mathbb{R}^3 \mapsto \mathbb{R}$ and a differentiable Marching Tetrahedral layer ${DMT}(\cdot)$ to transfer an implicit SDF into an explicit watertight mesh $M_{wt}$.
This process is formulated as: $M_{wt} = {DMT}(s(q))$, where $q$ is a set of predefined sampled points.
By integrating the strengths of both implicit and explicit 3D representations (\textit{i.e.}, stable optimization properties and explicit structure), DMTet enables high-fidelity geometric modeling.

\textit{G-Shell}~\cite{gshell} is an expressive representation capable of modeling both watertight and non-watertight surfaces.
Building upon DMTet, G-Shell introduces a manifold signed distance field (mSDF) $\hat{s}: \mathbb{R}^3 \mapsto \mathbb{R}$ on a watertight template. 
In this field, the sign indicates whether a point lies on an open surface, while the absolute value represents the geodesic distance to the boundary.
A non-watertight mesh $M_{nwt}$ can then be extracted through the G-Shell mesh extraction process ${GSE}(\cdot)$: $M_{nwt} = {GSE}(s(q), \hat{s}(q))$.
Additionally, G-Shell can also model a watertight mesh $M_{wt}$ by ignoring mSDF. 
In this case, G-Shell is equivalent to DMTet: $M_{wt} = {GSE}(s(q))$.

\subsection{Overview}
\label{sec:3.2}
\begin{figure*}[tp]
    \centering
    \includegraphics[width=0.9\textwidth]{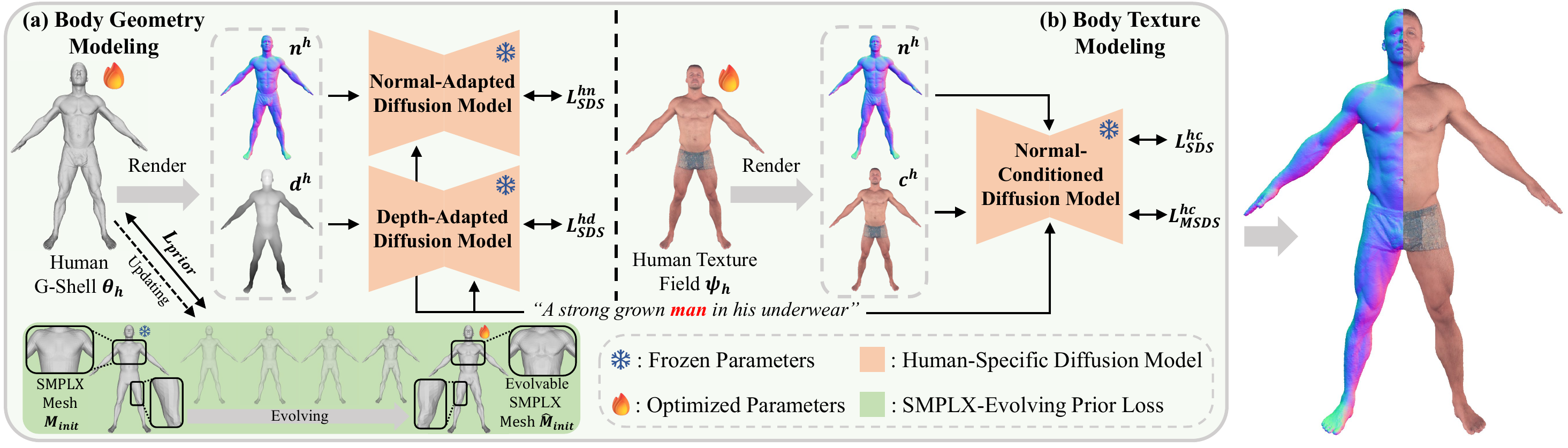} 
    \caption{The process for generating a basic human model involves two steps:
    \textbf{(a)} Employing human-specific geometry-aware diffusion models and the SMPLX-evolving prior loss to model realistic and reasonable body shapes.
    \textbf{(b)} Subsequently, using a normal-conditioned diffusion model to generate lifelike human textures.}
    \label{fig:pipeline1}
    \vspace{-0.5cm}
\end{figure*}

\begin{figure*}[tp]
    \centering
    \includegraphics[width=0.9\textwidth]{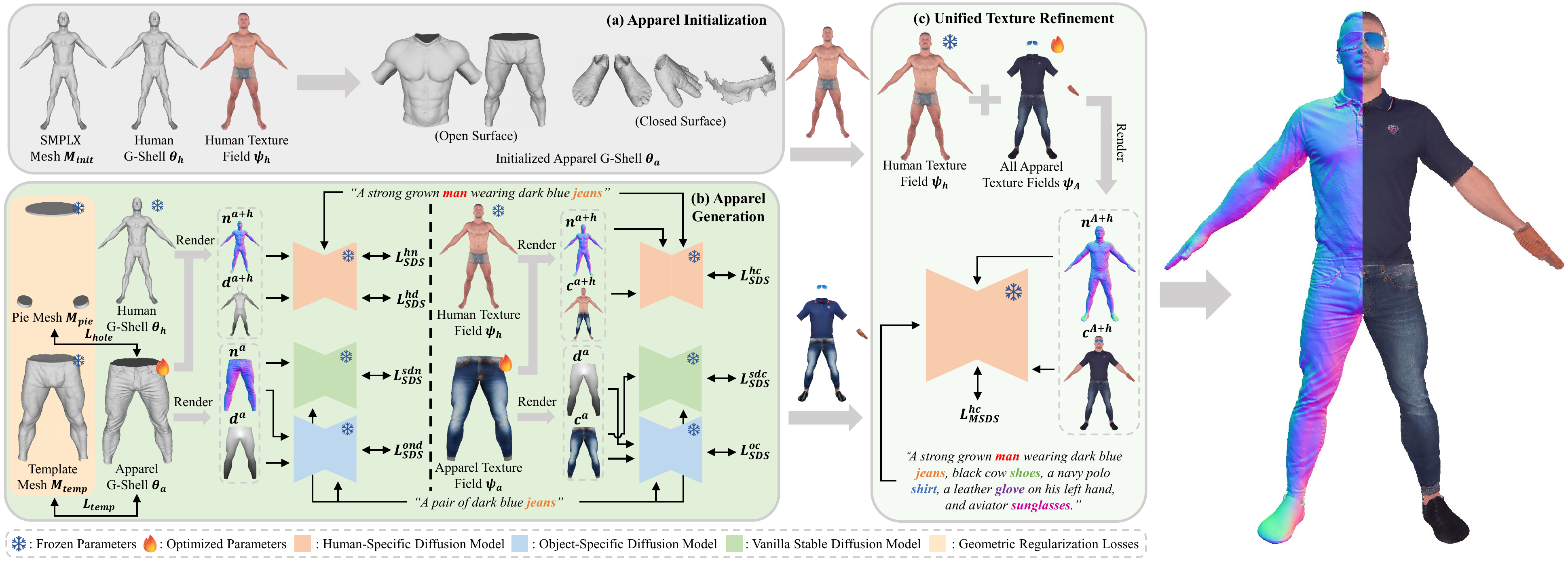} 
    \caption{The process for generating apparel involves three steps:
    \textbf{(a)} Initializing apparel with the semantic-aligned human body.
    \textbf{(b)} Modeling apparel piece by piece using object-specific generative priors and geometric losses.
    \textbf{(c)} Refining the texture of the assembled avatar using a unified texture refinement process.}
    \label{fig:pipeline2}
    \vspace{-0.5cm}
\end{figure*}

\noindent
Given a text prompt, \blue{DreamBarbie} aims to create an animatable, disentangled 3D avatar dressed in simulation-ready garments, along with diverse shoes and accessories, resembling iconic Barbie dolls.
Specifically, our framework consists of three stages:
\textbf{(1) Human Body Generation}: This stage generates a reasonable and realistic basic human body by leveraging human-specific generative priors and a novel SMPLX-evolving prior loss (Sec.~\ref{sec:3.3} and Fig.~\ref{fig:pipeline1});
\textbf{(2)} \textbf{Apparel Generation}: This stage models high-quality garments, shoes, and accessories piece by piece, utilizing object-specific diffusion models together with several initialization strategies and geometric losses (Sec.~\ref{sec:3.4} and Fig.~\ref{fig:pipeline2});
and \textbf{(3) Unified Texture Refinement}: This stage enhances visual harmony and consistency by jointly fine-tuning the composed avatar (Sec.~\ref{sec:3.5} and Fig.~\ref{fig:pipeline2}).

\subsection{Human Body Generation}
\label{sec:3.3}
\noindent
Since the human body is a closed surface, we utilize G-Shell ${\theta}_{h}$ in watertight mesh mode (\textit{i.e.}, ignoring mSDF and retaining only SDF) and a texture field ${\psi}_{h}$ to model the geometry and appearance of the human body, respectively.
These components are optimized under human-specific generative priors and a novel SMPLX-evolving prior loss, as illustrated in Fig.~\ref{fig:pipeline1}.

\textit{Human Body Geometry Modeling.}
To enhance the robustness of 3D human generation, we first optimize the shape parameters $\beta$ of an SMPL-X model using SDS loss, aligning the coarse body shape with the input text description.
The resulting initial mesh, denoted $M_{init}$, serves two key purposes: (1) initializing the human G-Shell $\theta_h$, and (2) providing rich, semantically aligned human priors to guide realistic outfit generation.
As discussed in the introduction, using a general T2I model struggles to provide domain-specific constraints necessary for creating realistic human bodies or outfits.
To address this limitation, we employ human-specific diffusion models from HumanNorm~\cite{humannorm}, which are fine-tuned with high-fidelity human data.

In particular, the human-specific diffusion models include a normal-adapted diffusion model ${\phi}_{hn}$, a depth-adapted diffusion model ${\phi}_{hd}$ for human shape generation, and a normal-conditioned diffusion model ${\phi}_{hc}$ for human texture creation.
The geometry-aware diffusion models optimize the initialized human G-Shell ${\theta}_{h}$ using the following SDS losses:
\begin{align}
    \nabla_{{\theta}_{h}} \mathcal{L}^{{hn}}_{SDS} &= \mathbb{E}_{t, \epsilon} \left[( {\epsilon}_{{\phi}_{hn}}(n^{h}_t; y_{h}, t) - \epsilon ) \frac{\partial n^{h}}{\partial {\theta}_{h}} \right],
    \label{eq:sds_human_normal} \\
    \nabla_{{\theta}_{h}} \mathcal{L}^{{hd}}_{SDS} &= \mathbb{E}_{t, \epsilon} \left[( {\epsilon}_{{\phi}_{hd}}(d^{h}_t; y_{h}, t) - \epsilon ) \frac{\partial d^{h}}{\partial {\theta}_{h}} \right],
    \label{eq:sds_human_dpeth}
\end{align}
where $n^{h}$ and $d^{h}$ are the rendered normal and depth maps of the human body, respectively.
$y_{h}$ denotes the input minimal-clothed human body description.

\textit{SMPLX-Evolving Prior Loss.}
Although Eq.~\ref{eq:sds_human_normal} and Eq.~\ref{eq:sds_human_dpeth} enable the creation of intricate human bodies, the overly strong human-specific generative priors cause the generated results to overfit the input text. This leads to unnatural geometry and exaggerated body proportions (Fig.~\ref{fig:ablat}-(a)).
A straightforward approach is to introduce parametric human body models (\textit{e.g.}, SMPL-X) and provide human body priors using the following equation:
\begin{equation}
    \mathcal{L}_{{prior}} = \sum_{p \in P} \left\| s_{\theta_{h}}(p) - s_{{init}}(p) \right\|_2^2,
    \label{eq:prior}
\end{equation}
where $s_{\theta_{h}}(\cdot)$ and $s_{{init}}(\cdot)$ represent the SDF of the generated body G-Shell ${\theta}_{h}$ and the initialized SMPL-X mesh $M_{init}$, respectively, and $P$ is a set of randomly sampled points in space.
However, due to the overly smooth geometry of SMPL-X, this approach may lead to a reduction in diversity and fine details (Fig.~\ref{fig:ablat}-(a)).

To address this problem, we propose the SMPLX-evolving prior loss inspired by the evolving constraint introduced in SeeAvatar~\cite{seeavatar}.
Specifically, we freeze all parameters of the SMPL-X model and enhance the initial mesh $M_{init}$  into an evolvable representation $\hat{M}_{init}$ by adding learnable vertex-wise offsets.
These offsets are periodically fitted to the current body every $\delta$ iteration, allowing the optimized offsets to model complex body features.
The fitting objective is as follows:
\begin{equation}
    \mathcal{L}_{fit} = \lambda_{{chamf}} \mathcal{L}_{chamf} + \lambda_{{edge}} \mathcal{L}_{edge} + \lambda_{{nor}} \mathcal{L}_{nor} + \lambda_{{lap}} \mathcal{L}_{lap},
\end{equation}
where $\mathcal{L}_{chamf}$ is the Chamfer distance between $\hat{M}_{init}$ and the current body shape, $\mathcal{L}_{edge}$, $\mathcal{L}_{nor}$ and $\mathcal{L}_{lap}$ are the edge length regularization, normal consistency loss, and Laplacian smoothness, respectively.
The advantages of this loss are twofold: 
\textbf{(1) Enhanced Modeling Capabilities}: Compared to the smooth $M_{init}$, our $\hat{M}_{init}$ excels at capturing detailed geometric features, such as muscle contours (see the zoomed area at the bottom of Fig.~\ref{fig:pipeline1}), and offers reliable yet diverse priors for subsequent geometry generation (Fig.~\ref{fig:ablat}-(a)).
\textbf{(2) Topology Preservation}: Unlike the evolving templates with arbitrary topology in SeeAvatar~\cite{seeavatar}, our $\hat{M}_{init}$ preserves the SMPL-X model's topology and semantics during the evolution process. 
This makes it particularly suitable for apparel initialization, composition, animation, and simulation (Fig.~\ref{fig:app}).

In summary, the loss function for optimizing the human body geometry is formulated as follows:
\begin{equation}
    \mathcal{L}_{{hum-geo}} = \mathcal{L}^{{hn}}_{SDS} + \mathcal{L}^{{hd}}_{SDS} + \lambda_{{prior}} \mathcal{L}_{{prior}}.
    \label{eq:human_geometry}
\end{equation}

\textit{Human Body Texture Modeling.}
Given the human mesh generated from the previous stage, we fix it and utilize a texture field ${\psi}_{h}$ which maps a query position to its color to generate a normal-aligned body appearance.
This field is optimized using a normal-conditioned diffusion model ${\phi}_{hc}$ with loss defined as:
\begin{equation}
    \nabla_{{\psi}_{h}} \mathcal{L}^{{hc}}_{SDS} = \mathbb{E}_{t, \epsilon} \left[( {\epsilon}_{{\phi}_{hc}}(c^{h}_t; n^{h}, y_{h}, t) - \epsilon ) \frac{\partial c^{h}}{\partial {\psi}_{h}} \right],
    \label{eq:sds_human_color}
\end{equation}
where $c^{h}$ represents the rendered color image of the generated human body.
Since the vanilla SDS loss often leads to color oversaturation, we replace it with the following multi-step SDS (MSDS) loss~\cite{humannorm} to further enhance the texture's realism in later iterations of texture optimization:
\begin{equation}
    \begin{aligned}
    &\nabla_{{\psi}_{h}} \mathcal{L}^{{hc}}_{MSDS} = \mathbb{E}_{t, \epsilon} \left[( h(c^{h}_t; n^{h}, y_{h}, t) - \epsilon ) \frac{\partial c^{h}}{\partial {\psi}_{h}} \right] + \\ & \mathbb{E}_{t, \epsilon} \left[( V(h(c^{h}_t; n^{h}, y_{h}, t)) - V(\epsilon) ) \frac{\partial V(\epsilon)}{\partial \epsilon} \frac{\partial c^{h}}{\partial {\psi}_{h}} \right],
    \label{eq:msds_human_color}
    \end{aligned}
\end{equation}
where $V$ denotes the first $k$ layers of the VGG network~\cite{vgg}, and $h(\cdot)$ represents the multi-step image generation function of the normal-aligned diffusion model.

\subsection{Apparel Generation}
\label{sec:3.4}
\noindent
Given an SMPLX-aligned basic body generated in the previous stage, we proceed to dress the Barbie-like avatar in various outfits.
Similar to body modeling, we employ G-Shell ${\theta}_{a}$ for apparel geometry and a texture field ${\psi}_{a}$ for apparel texture.
Closed shoes and accessories are represented using G-Shell in watertight mesh mode (\textit{i.e.}, ignoring mSDF and retaining only SDF), while open garments are modeled using G-Shell in non-watertight mesh mode (\textit{i.e.}, retaining both SDF and mSDF).
As illustrated in Fig.~\ref{fig:pipeline2}, the apparel generation process begins by initializing apparel using predefined SMPL-X masks that cover over a dozen daily outfits.
We then create high-quality apparel piece by piece using object-specific generative priors and well-designed geometric losses.
Lastly, we fine-tune the assembled avatar to improve appearance harmony and consistency.

\begin{figure}[tp]
    \centering
    \includegraphics[width=0.45\textwidth]{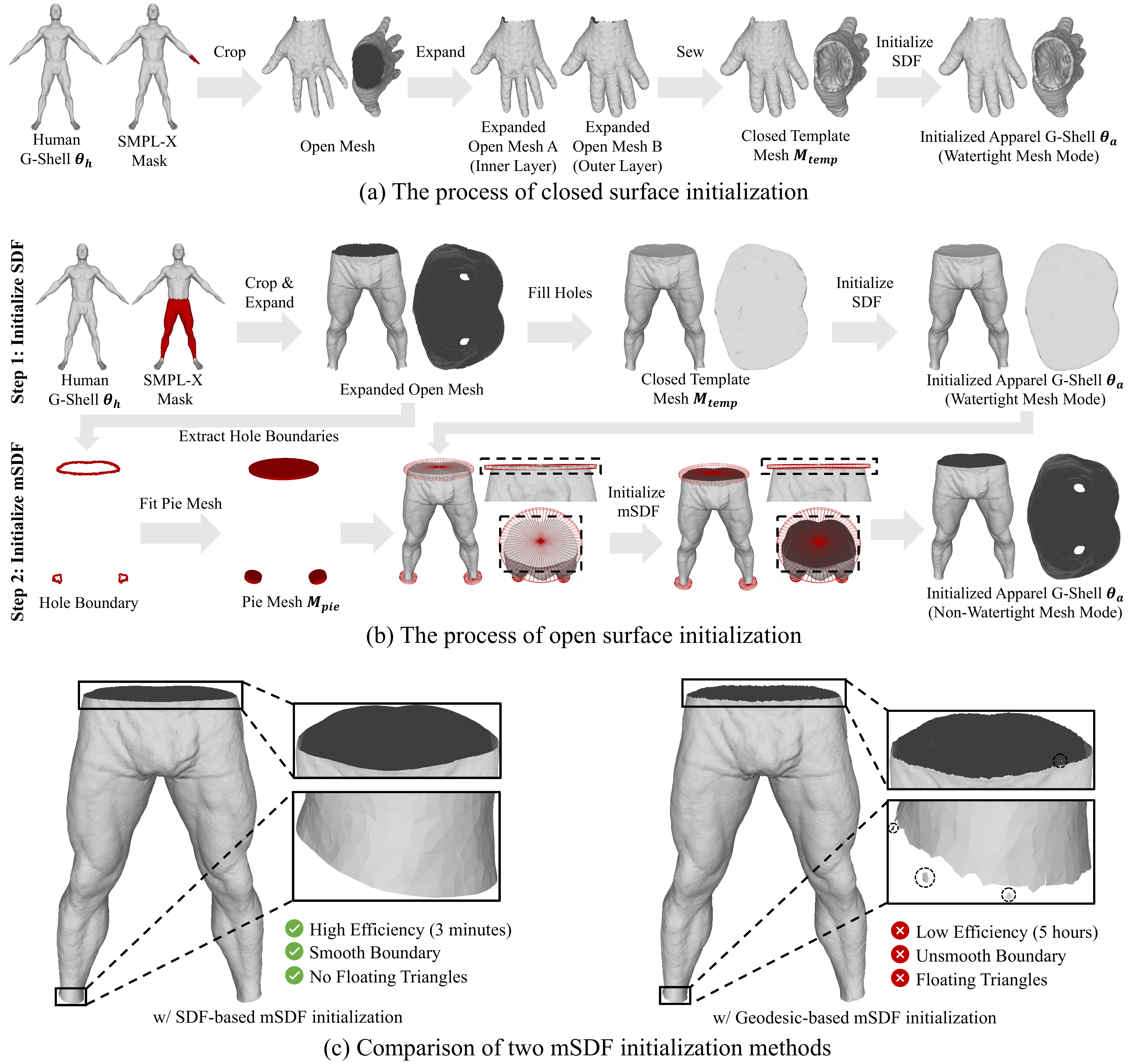} 
    \caption{\textbf{(a) Closed Surface Initialization}: Expand and sew the open mesh (cropped via SMPL-X mask) to create a closed template mesh $M_{temp}$, used to initialize the SDF $s_{\theta_{a}}(\cdot)$ of ${\theta}_{a}$.
    \textbf{(b) Open Surface Initialization}: Fit a watertight pie mesh $M_{pie}$ over the holes of the cropped open mesh. Use its SDF values to initialize the mSDF $\hat{s}_{\theta_{a}}(\cdot)$ of ${\theta}_{a}$.
    \textbf{(c) Comparison}: Contrast geodesic-based and SDF-based mSDF initialization.}
    \label{fig:init}
    \vspace{-0.5cm}
\end{figure}

\textit{Apparel Initialization.}
In the absence of multi-view supervision~\cite{gshell,d3human}, text-driven optimization of G-Shell representations heavily depends on effective initialization, especially for open surface modeling.
Therefore, we carefully design two strategies for initializing open clothes and closed shoes/accessories, enabling accurate topology modeling for each category.
\textbf{(1) Closed Surface Initialization}:
We crop the open sub-mesh of the human body using the corresponding SMPL-X mask, expand it along vertex normals to create inner and outer layers, and sew them to form a double-layer closed template mesh $M_{temp}$.
This mesh initializes the SDF $s_{\theta_{a}}(\cdot)$ of the apparel G-Shell ${\theta}_{a}$.
Fig.~\ref{fig:init}-(a) shows this process.
\blue{
\textbf{(2) Open Surface Initialization}:
As mSDF requires a closed manifold domain, we first construct a single-layer watertight template, $M_{temp}$, by hole-filling the cropped body mesh.
This simplified topology serves as the canonical domain for the mSDF. 
The primary challenge lies in defining the open boundaries on $M_{temp}$.
While a geodesic-based initialization (\textit{i.e.}, signed geodesic distance to boundaries) is intuitive, it is computationally prohibitive ($\sim$5 hours) and mathematically ill-posed for 3D implicit fields.
Because geodesics are confined to the 2-manifold, they lack a stable gradient flow in 3D space, resulting in optimization artifacts (Fig.~\ref{fig:init}-(c)).}

\blue{
To address this, we reformulate boundary modeling as a field-based intersection problem.
Specifically, we introduce the Pie Mesh ($M_{pie}$), a localized watertight proxy that encapsulates each hole's boundary (see Supp. Mat. for fitting details). 
Mathematically, instead of manifold traversal, we use $M_{pie}$'s SDF values to initialize the mSDF $\hat{s}_{\theta_{a}}(\cdot)$.
This ensures that points sampled inside $M_{pie}$ have negative mSDF values (indicating they lie inside the open surface), while those outside have positive values (indicating they lie outside the open surface). 
Consequently, the open surface is implicitly determined by the set subtraction between the $M_{temp}$ manifold and the $M_{pie}$ volume (Fig.~\ref{fig:init}-(b), black dotted box).
Fig.~\ref{fig:init}-(b) illustrates the whole process.
This strategy offers three advantages:
\textbf{(1) Efficiency}: Replacing manifold-constrained path-finding with SDF queries achieves a $100\times$ speedup ($\sim$3 mins vs. $\sim$5 hours, see the Supp. Mat. for speed comparison details).
\textbf{(2) Stability}: The SDF provides a globally continuous gradient field, ensuring stable supervision and eliminating floating artifacts or unsmooth boundaries.
\textbf{(3) Regularization}: $M_{pie}$ serves as a geometric prior, providing explicit boundary constraints for subsequent optimization (see Eq.~\ref{eq:hole}).
Fig.~\ref{fig:init}-(c) compares these two initialization methods.}

\textit{Apparel Geometry Modeling.}
Similar to human body geometry generation, we utilize human-specific diffusion models to optimize outfit geometry with the SDS loss: 
\begin{align}
    \nabla_{{\theta}_{a}} \mathcal{L}^{{hn}}_{SDS} &= \mathbb{E}_{t, \epsilon} \left[( {\epsilon}_{{\phi}_{hn}}(n^{a+h}_t; y_{a+h}, t) - \epsilon ) \frac{\partial n^{a+h}}{\partial {\theta}_{a}} \right],
    \label{eq:sds_apparel_normal} \\
    \nabla_{{\theta}_{a}} \mathcal{L}^{{hd}}_{SDS} &= \mathbb{E}_{t, \epsilon} \left[( {\epsilon}_{{\phi}_{hd}}(d^{a+h}_t; y_{a+h}, t) - \epsilon ) \frac{\partial d^{a+h}}{\partial {\theta}_{a}} \right],
    \label{eq:sds_apparel_dpeth}
\end{align}
where $y_{a+h}$ is the text description of the clothed avatar, and $n^{a+h}$ and $d^{a+h}$ are the rendered clothed human normal and depth maps, respectively.
However, human-specific diffusion models are not well-suited for creating high-quality clothes, shoes, and accessories because they are not good at modeling general 3D objects.
Hence, we additionally introduce the object-specific diffusion models~\cite{richdreamer} pretrained on the LAION dataset~\cite{laion} for providing in-domain details and diversity.

Concretely, these diffusion models include a normal-depth diffusion model ${\phi}_{ond}$ for optimizing the apparel geometry, and a depth-conditional diffusion model ${\phi}_{oc}$ for creating the apparel texture. 
The normal-depth model provides effective supervision for outfit geometry generation by accurately modeling the joint distribution of normal and depth maps using the following SDS loss: 
\begin{equation}
    \begin{aligned}
    \nabla_{{\theta}_{a}} \mathcal{L}^{{ond}}_{SDS} &= \mathbb{E}_{t, \epsilon} \left[( {\epsilon}_{{\phi}_{ond}}(n^{a}_t; y_{a}, t) - \epsilon ) \frac{\partial n^{a}}{\partial {\theta}_{a}} \right] \\ 
    &+ \mathbb{E}_{t, \epsilon} \left[( {\epsilon}_{{\phi}_{ond}}(d^{a}_t; y_{a}, t) - \epsilon ) \frac{\partial d^{a}}{\partial {\theta}_{a}} \right],
    \label{eq:sds_apparel_normal_depth}
    \end{aligned}
\end{equation}
\begin{equation}
    \nabla_{{\theta}_{a}} \mathcal{L}^{{sdn}}_{SDS} = \mathbb{E}_{t, \epsilon} \left[( {\epsilon}_{{\phi}_{sd}}(n^{a}_t; y_{a}, t) - \epsilon ) \frac{\partial n^{a}}{\partial {\theta}_{a}} \right],
    \label{eq:van_sds_apparel_normal}
\end{equation}
where $y_{a}$ is the description of a single outfit, and $n^{a}$ and $d^{a}$ are the rendered normal and depth maps of the apparel in the undressed state.
Besides, $\mathcal{L}^{{sdn}}_{SDS}$ is the vanilla Stable Diffusion (SD) SDS loss enforced on the rendered apparel normal maps.
As shown in RichDreamer~\cite{richdreamer}, naive SD helps object-specific diffusion models produce more stable results.
In this way, two specific generative priors provide powerful and reasonable guidance to modeling high-fidelity outfits (Fig.~\ref{fig:ablat}-(b)).

\begin{figure*}[tp]
    \centering
    \includegraphics[width=0.9\textwidth]{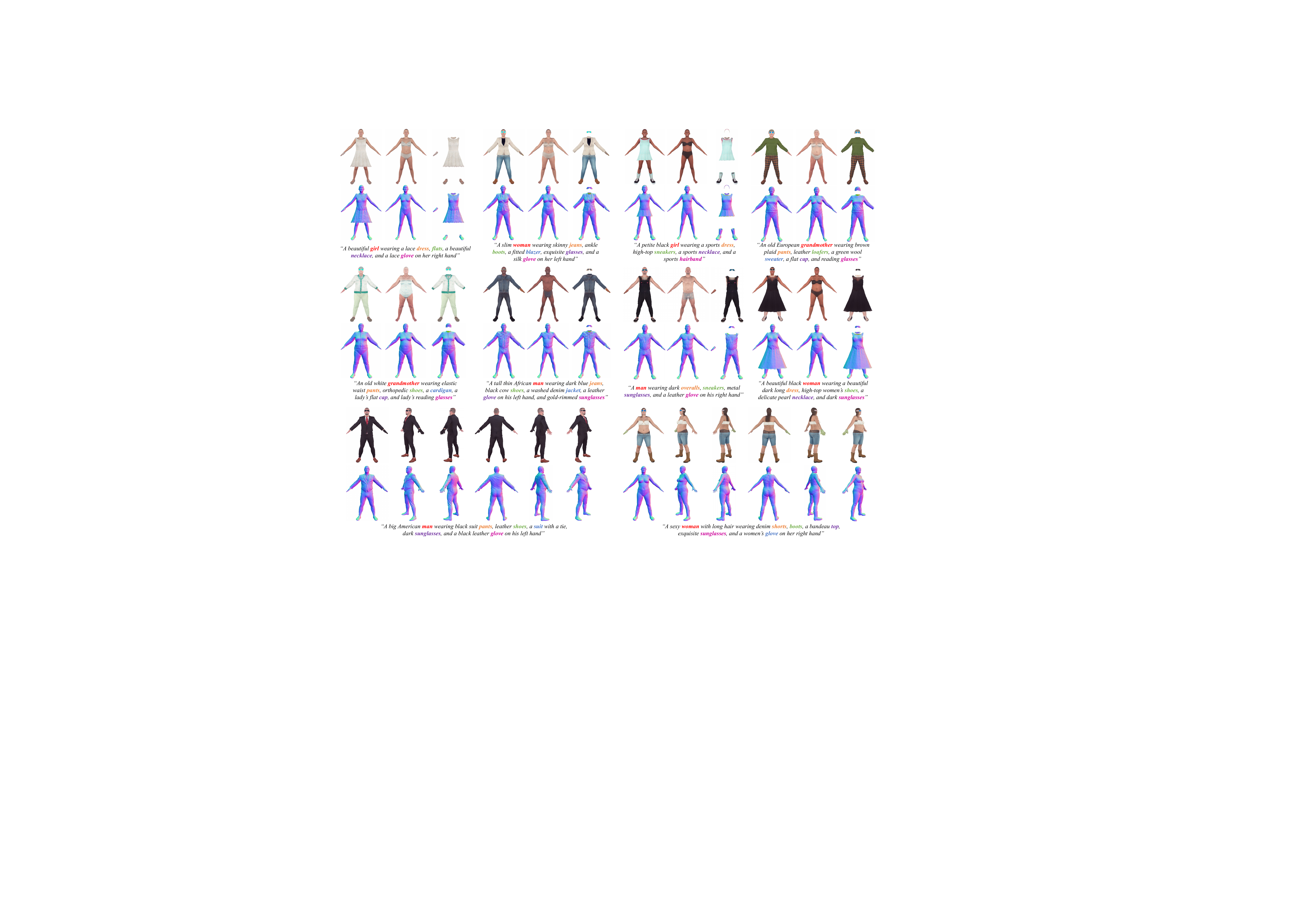} 
    \caption{Diverse Range of \textbf{Barbie-Style} Avatar Generation. Rendering color images and normal images for visualization. \textbf{Please zoom in to see the details and see Supp. Mat. for video results.}}
    \label{fig:show}
    \vspace{-0.5cm}
\end{figure*}

\textit{Template-Preserving Loss.}
Similar to human-specific diffusion models, relying on object-specific diffusion models may lead to geometric artifacts (\textit{e.g.}, unexpected holes, see Fig.~\ref{fig:ablat}-(d)).
To address this problem, we propose a template-preserving loss that enforces the apparel geometry to fit the template mesh $M_{temp}$, ensuring geometric integrity. This loss is formulated as follows:
\begin{equation}
    \mathcal{L}_{{temp}} = \sum_{p \in P} \left\| s_{\theta_{a}}(p) - s_{{temp}}(p) \right\|_2^2,
    \label{eq:template}
\end{equation}
where $s_{temp}(\cdot)$ represents the SDF of the template mesh $M_{temp}$, and $P$ is the set of randomly sampled points in space.

\textit{Hole-Preserving Loss.}
Since the object-specific diffusion model is primarily trained on watertight mesh data, it tends to guide the G-Shell to model closed surfaces, resulting in many floating triangles inside the holes (Fig.~\ref{fig:ablat}-(e)).
Moreover, due to the absence of multi-image input, we cannot reliably supervise the open surface like other G-Shell reconstruction works~\cite{gshell,d3human}.
To address this issue, we propose a hole-preserving loss that exploits the pie mesh $M_{pie}$ obtained from the SDF-based mSDF initialization.
This loss is formulated as:
\begin{equation}
    \mathcal{L}_{{hole}} = \sum_{p \in P} \left\| \hat{s}_{\theta_{a}}(p) - s_{{pie}}(p) \right\|_2^2,
    \label{eq:hole}
\end{equation}
where $s_{{pie}}(\cdot)$ represents the SDF of the pie mesh $M_{pie}$.
This loss effectively suppresses floating triangles by using the SDF of $M_{pie}$ to supervise the mSDF of ${\theta}_{a}$, thereby preserving clean and well-defined hole structures.

\textit{Collision Loss.}
To ensure the generated apparel does not intersect with the underlying human body (Fig.~\ref{fig:ablat}-(f)), we introduce the following collision loss $\mathcal{L}_{{colli}}$:
\begin{equation}
    \mathcal{L}_{{colli}} = \sum_{v \in V} \text{relu}(-s_{h}(v)),
    \label{eq:collision}
\end{equation}
where $V$ is the set of vertices of the generated apparel mesh, and $s_{h}(\cdot)$ is the SDF of the human body.

In summary, the loss function for optimizing the apparel geometry is as follows:
\begin{equation}
    \begin{aligned}
        & \mathcal{L}_{{app-geo}} = \mathcal{L}^{{hn}}_{SDS} + \mathcal{L}^{{hd}}_{SDS} + \mathcal{L}^{{ond}}_{SDS} + \mathcal{L}^{{sdn}}_{SDS} \\ & + \lambda_{{temp}} \mathcal{L}_{{temp}} + \lambda_{{hole}} \mathcal{L}_{{hole}} + \lambda_{{colli}} \mathcal{L}_{{colli}},
    \label{eq:apparel_geometry}
    \end{aligned}
\end{equation}
where $\mathcal{L}_{{hole}}$ is only used in non-watertight mesh mode.

\textit{Apparel Texture Modeling.}
Given the generated apparel shape and the textured human body, we employ an object-specific depth-conditional diffusion model for lifelike apparel textures with the following SDS loss:
\begin{equation}
    \nabla_{{\psi}_{a}} \mathcal{L}^{{oc}}_{SDS} = \mathbb{E}_{t, \epsilon} \left[( {\epsilon}_{{\phi}_{oc}}(c^{a}_t; d^{a}, y_{a}, t) - \epsilon ) \frac{\partial c^{a}}{\partial {\psi}_{a}} \right],
    \label{eq:sds_apparel_depth_color}
\end{equation}
\begin{equation}
    \nabla_{{\psi}_{a}} \mathcal{L}^{{sdc}}_{SDS} = \mathbb{E}_{t, \epsilon} \left[( {\epsilon}_{{\phi}_{sd}}(c^{a}_t; y_{a}, t) - \epsilon ) \frac{\partial c^{a}}{\partial {\psi}_{a}} \right],
    \label{eq:van_sds_apparel_color}
\end{equation}
where $c^{a}$ is the color image. 
Additionally, human-specific diffusion models are combined to optimize the appearance of outfits in the dressed state to ensure basic texture harmony, and the SDS loss is formulated as:
\begin{equation}
    \nabla_{{\psi}_{a}} \mathcal{L}^{{hc}}_{SDS} = \mathbb{E}_{t, \epsilon} \left[( {\epsilon}_{{\phi}_{hc}}(c^{a+h}_t; n^{a+h}, y_{a+h}, t) - \epsilon ) \frac{\partial c^{a+h}}{\partial {\psi}_{a}} \right].
    \label{eq:sds_apparel_normal_color}
\end{equation}

In summary, the loss function for optimizing the apparel appearance is as follows:
\begin{equation}
    \mathcal{L}_{{app-tex}} = \mathcal{L}^{{hc}}_{SDS} + \mathcal{L}^{{oc}}_{SDS} + \mathcal{L}^{{sdc}}_{SDS}.
    \label{eq:apparel_texture}
\end{equation}

\subsection{Unified Texture Refinement}
\label{sec:3.5}
\noindent
After the above steps, we produce detailed geometry and lifelike textures for human bodies, garments, shoes, and accessories.
However, a domain gap in the training data for fine-tuning expert models causes texture disharmony between the body and outfits, reducing realism (Fig.~\ref{fig:ablat}-(c)).
To address this issue, we propose a unified texture refinement (UTR) strategy.
Specifically, we fine-tune all apparel texture fields ${\psi}_{A} = \{ \psi_{{a}_i}, i \in [1, ..., N] \}$ of the assembled avatar under the human-specific normal-conditioned diffusion model and the MSDS loss from Eq.~\ref{eq:msds_human_color}.
The UTR strategy ensures visual unity and improves texture harmony across the avatar.
It reduces the domain gap, enhancing realism and aesthetic quality.

\subsection{Implementation Details}
\label{sec:3.6}
\noindent
Our algorithm is implemented using PyTorch~\cite{pytorch} and ThreeStudio~\cite{threestudio}.
\blue{Generating a human body takes $\sim$3 hours (24GB VRAM); generating an outfit takes $\sim$4 hours (40GB VRAM).}
We use MLP with multi-resolution hash encoding~\cite{instant-ngp} to efficiently represent SDF, mSDF, and color fields.
Given a text describing a dressed human, \textit{e.g.}, ``A man wearing jeans and a T-shirt'', we first use a rule-based script to decompose the full input text into multiple sub-prompts: ``A man in his underwear'', ``A pair of jeans'', and ``A T-shirt''.
We then generate a base human body based on ``A man in his underwear''.
Next, we generate jeans based on ``A pair of jeans'' and ``A man wearing jeans'', followed by generating the T-shirt based on ``A T-shirt'' and ``A man wearing jeans and a T-shirt''.
Finally, fine-tune the texture of the assembled avatar based on the full input text ``A man wearing jeans and a T-shirt''.
Additional details can be found in the Supp. Mat.

\begin{figure*}[tp]
    \centering
    \includegraphics[width=0.9\textwidth]{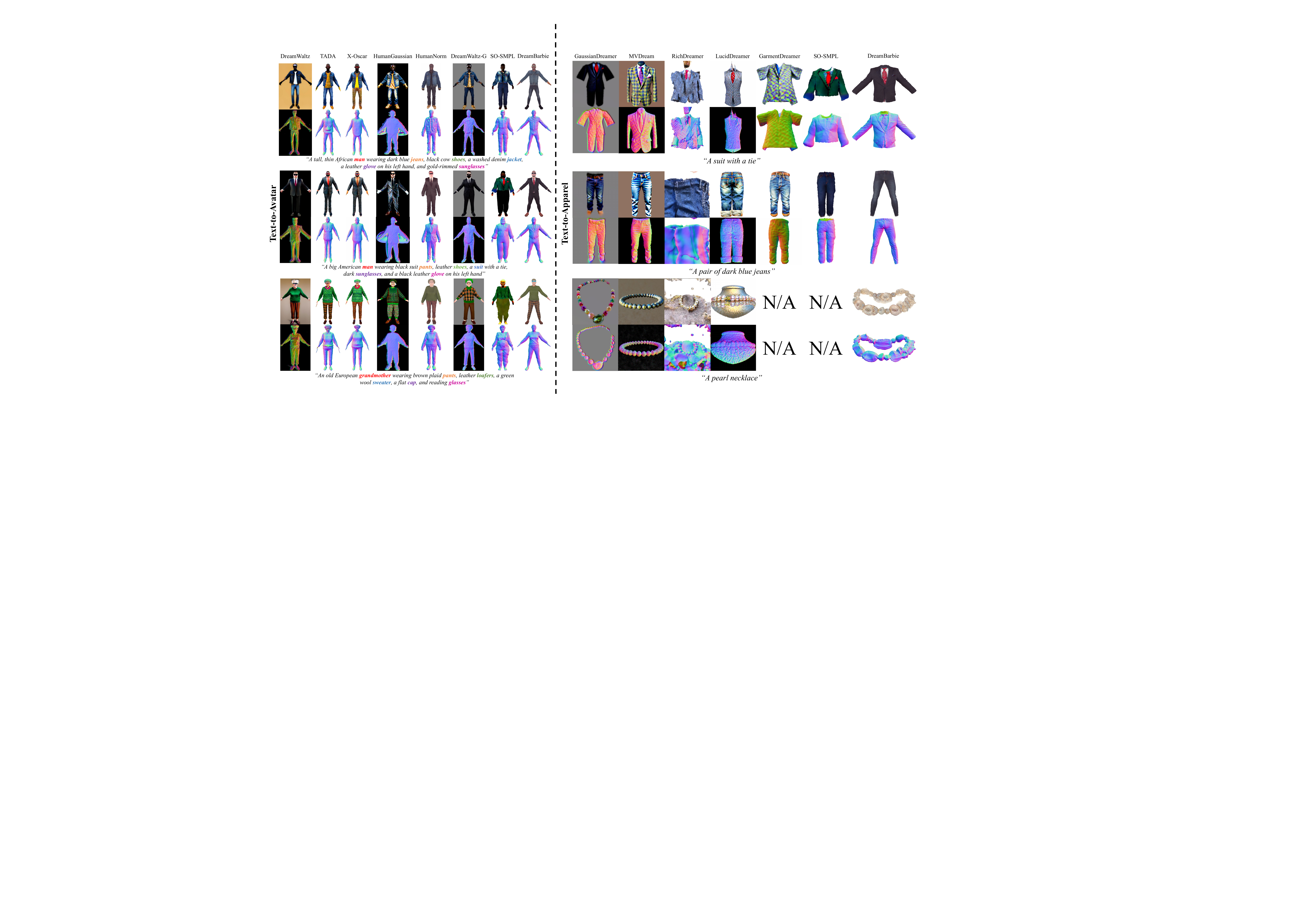} 
    \caption{Qualitative comparisons with baseline text-to-avatar and text-to-apparel methods. GarmentDreamer~\cite{garmentdreamer} and SO-SMPL~\cite{so-smpl} only support clothing generation.}
    \label{fig:comp1}
    \vspace{-0.5cm}
\end{figure*}

\begin{figure}[tp]
    \centering
    \includegraphics[width=0.45\textwidth]{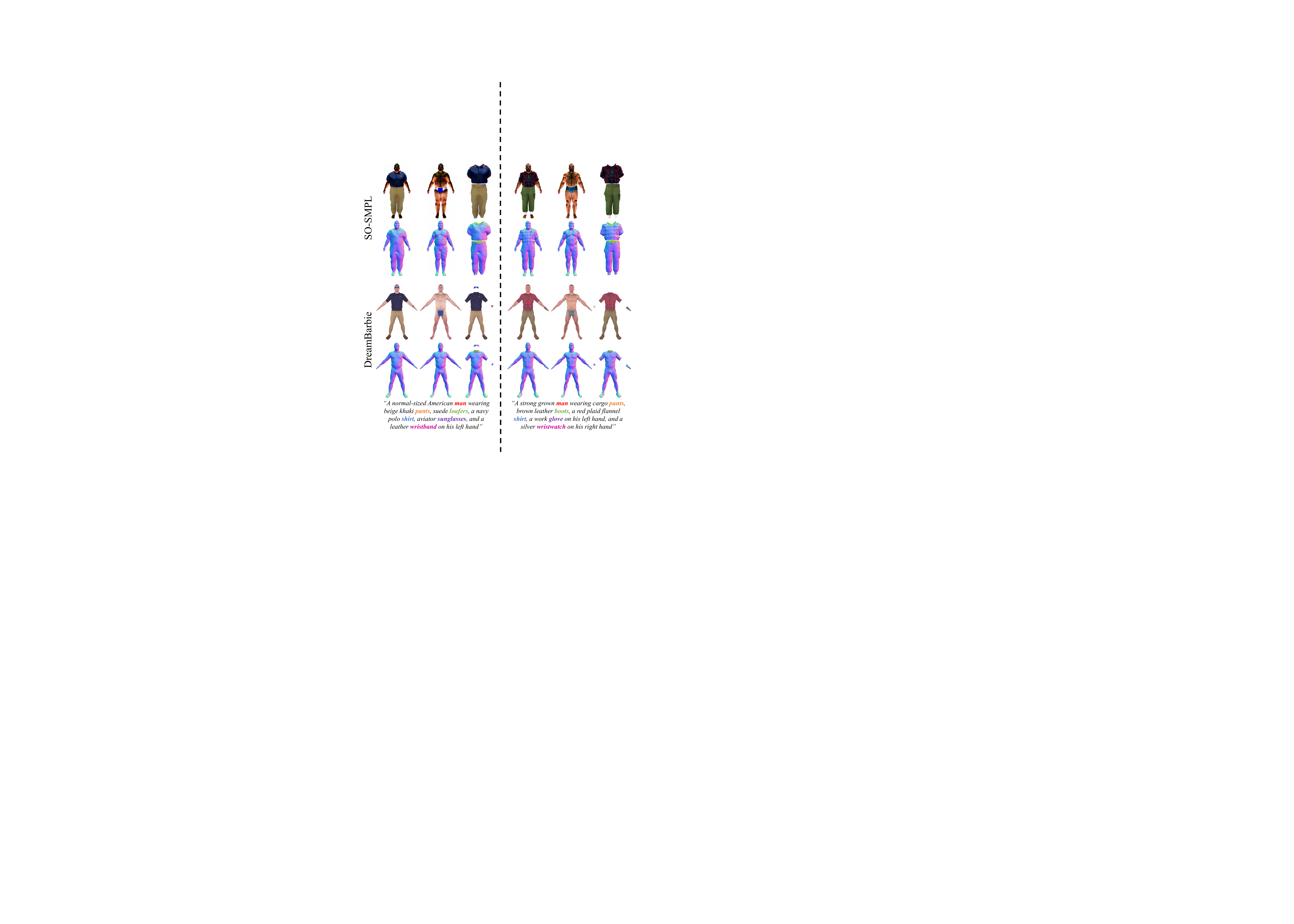} 
    \caption{Qualitative comparisons with the text-to-disentangled-avatar work SO-SMPL~\cite{so-smpl}. 
    SO-SMPL not only exhibits significantly lower generation quality than our method but also fails to produce the diverse outfits shown in Fig.~\ref{fig:show}, as achieved by our Barbie-style avatars.}
    \label{fig:comp2}
    \vspace{-0.5cm}
\end{figure}

\vspace{-0.5cm}
\section{Experiments}
\label{sec:4}

\noindent
Some examples of 3D bodies and outfits generated by our method are shown in Fig.~\ref{fig:show}. 
Due to space limitations, only the key experimental settings and results are presented here. 
For complete results and details, please refer to our Supp. Mat.

\subsection{Experimental Settings}
\label{sec:4.1}
\noindent
\textit{Baselines.}
We perform both quantitative and qualitative comparisons of \blue{DreamBarbie} with state-of-the-art (SOTA) SDS-based methods for text-driven avatar generation and apparel generation.
For avatar generation, we compare with text-to-holistic-avatar methods (DreamWaltz~\cite{dreamwaltz}, TADA~\cite{tada}, X-Oscar~\cite{x-oscar}, HumanGaussian~\cite{humangaussian}, HumanNorm~\cite{humannorm}, and DreamWaltz-G~\cite{dreamwaltzg}) and text-to-decoupled-avatar method (SO-SMPL~\cite{so-smpl}).
For text-to-apparel generation, we compare with text-to-3D methods (MVDream~\cite{mvdream}, GaussianDreamer~\cite{gaussiandreamer}, RichDreamer~\cite{richdreamer}, LucidDreamer~\cite{luciddreamer}), text-to-garment method (GarmentDreamer~\cite{garmentdreamer}), and SO-SMPL~\cite{so-smpl}.

\begin{table}
    \begin{center}
    \caption{Quantitative comparisons with baseline methods}
    \label{tab:quantitative_comparisons}
    \resizebox{0.45\textwidth}{!}{
    \begin{threeparttable}
    \begin{tabular}{c|c|c|c|c}
    \hline
    Method & BLIP-VQA $\uparrow$ & BLIP2-VQA $\uparrow$ & GQP $\uparrow$ & TAP $\uparrow$ \\
    \hline
    DreamWaltz & 0.5819 & 0.5333 & \blue{2.80} & \blue{2.31} \\
    TADA & 0.5306 & 0.5333 & \blue{3.25} & \blue{2.91} \\
    X-Oscar & 0.5069 & 0.5292 & \blue{3.43} & \blue{3.62} \\
    HumanGaussian & 0.6222 & 0.5069 & \blue{2.35} & \blue{2.05} \\
    HumanNorm & 0.5611 & 0.5292 & \blue{3.81} & \blue{3.51} \\
    DreamWaltz-G & 0.6444 & 0.5722 & \blue{2.87} & \blue{2.54} \\
    SO-SMPL & 0.6500$^\dagger$ & 0.6000$^\dagger$ & \blue{0.45} & \blue{0.71} \\
    \blue{DreamBarbie} (Ours) & \textbf{0.7167}/\textbf{0.7333$^\dagger$} & \textbf{0.6333}/\textbf{0.7000$^\dagger$} & \textbf{\blue{81.04}} & \textbf{\blue{82.35}} \\
    \hline
    MVDream & 0.7000 & 0.5933 & \blue{9.03} & \blue{8.92} \\
    GaussianDreamer & 0.7017 & 0.5283 & \blue{3.84} & \blue{4.14} \\
    RichDreamer & 0.8100 & 0.6667 & \blue{3.25} & \blue{2.95} \\
    LucidDreamer & 0.7533 & 0.6400 & \blue{4.59} & \blue{3.77} \\
    GarmentDreamer & 0.6500$^\dagger$ & 0.4500$^\dagger$ & \blue{0.78} & \blue{0.63} \\
    SO-SMPL & 0.8500$^\dagger$ & 0.7667$^\dagger$ & \blue{0.26} & \blue{0.37} \\
    \blue{DreamBarbie} (Ours) & \textbf{0.8667}/\textbf{0.9667$^\dagger$} & \textbf{0.7667}/\textbf{0.9167$^\dagger$} & \textbf{\blue{78.25}} & \textbf{\blue{79.22}} \\
    \hline
    \end{tabular}
    \begin{tablenotes}
        \item The best result is highlighted in \textbf{bolded}. GQP: generation quality preference (\%), TAP: text-image alignment preference (\%). $\dagger$: When calculating VQA metrics, we ignore the problem of retrieving shoes and accessories, as GarmentDreamer~\cite{garmentdreamer} and SO-SMPL~\cite{so-smpl} only support clothing generation.
    \end{tablenotes}
    \end{threeparttable}
    }
    \end{center}
    \vspace{-0.8cm}
\end{table}

\textit{Dataset Construction.}
We utilized ChatGPT~\cite{chatgpt} to randomly generate 30 text descriptions of dressed avatars, with each example wearing a top, a bottom, a pair of shoes, and two random accessories. 
30 descriptions for dressed humans are used to evaluate avatar generation, and 30$\times$5 apparel descriptions are used to evaluate apparel generation (see the Supp. Mat. for the complete descriptions).

\begin{figure*}[tp]
    \centering
    \includegraphics[width=0.9\textwidth]{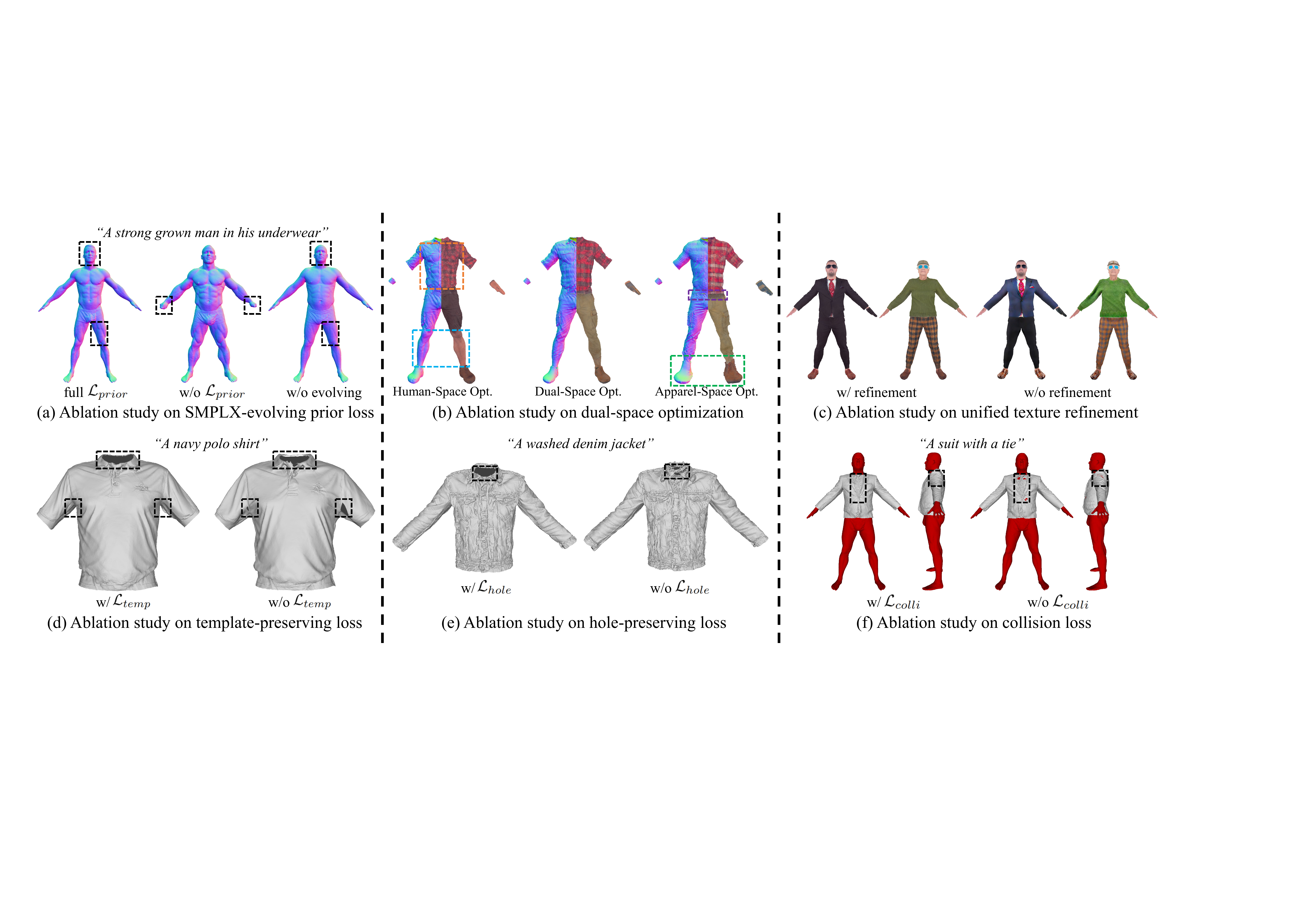} 
    \caption{Ablation Study of \blue{DreamBarbie}.}
    \label{fig:ablat}
    \vspace{-0.5cm}
\end{figure*}

\textit{Evaluation Metrics.}
Existing text-to-3D approaches utilize CLIP-based metrics to evaluate text-image alignment and compare the quality of generated results. 
However, CLIP-based metrics have been shown~\cite{t2i-compbench,llmscore} to be insufficient to accurately measure the fine-grained correspondence between 3D content and input prompts, which is further confirmed by experiments in the Supp. Mat.

Consequently, inspired by Progressive3D~\cite{progressive3d}, we adopt fine-grained text-to-image evaluation metrics, including BLIP-VQA~\cite{blip,lavis} and BLIP2-VQA~\cite{blip2,lavis}, to evaluate the generation capacity of current methods and \blue{DreamBarbie}. 
Specifically, we first convert the prompt into multiple separate questions to retrieve corresponding content, then feed the rendered image of the generated content into the VQA model and ask questions one by one, and finally use the probability of answering ``yes'' as the evaluation metric. 
For instance, the input avatar prompt ``A man wearing X1.'' is converted into ``Is the person in the picture a man?'' and ``Is the person in the picture wearing X1?''. 
The input apparel prompt ``A pair of X1.'' is converted into ``Is the object in the picture a pair of X1?''. 
\blue{Besides, we randomly select 40 cases (20 for avatar and 20 for outfit) to conduct a user study and ask 67 volunteers to assess (1) generation quality and (2) text-image alignment, and select the preferred methods.}

\subsection{Comparisons of Avatar Generation}
\label{sec:4.2}
\noindent
\textit{Comparison with Text-to-Avatar Methods.}
As shown in the upper part of Table~\ref{tab:quantitative_comparisons}, our method significantly outperforms all baseline methods across all evaluation metrics.
The qualitative results presented in the left part of Fig.~\ref{fig:comp1} further validate its superiority.
Compared to existing methods, \blue{DreamBarbie} generates avatars with more detailed and realistic geometry, while achieving better alignment with the input text, without omitting any specified apparel items.
Moreover, \blue{DreamBarbie} achieves a high degree of disentanglement among body, garments, shoes, and accessories, enabling flexible outfit combinations and edits, much like dressing physical Barbie dolls (Fig.~\ref{fig:app}-(a) and -(b)).
Furthermore, thanks to the use of G-Shell and the proposed SMPLX-evolving prior loss, \blue{DreamBarbie} supports expressive animation and compatibility with physical simulation (Fig.~\ref{fig:app}-(c) and -(d)).

\textit{Comparison with Text-to-Decoupled-Avatar Method.}
As illustrated in Fig.~\ref{fig:comp2}, our method surpasses SO-SMPL~\cite{so-smpl} in generating finer geometric details and more lifelike textures.
Additionally, \blue{DreamBarbie} enables the generation of various accessories (\textit{e.g.}, necklaces, glasses, and watches), which are not supported by SO-SMPL~\cite{so-smpl}.

\subsection{Comparisons of Apparel Generation}
\label{sec:4.3}
\begin{figure}[tp]
    \centering
    \includegraphics[width=0.45\textwidth]{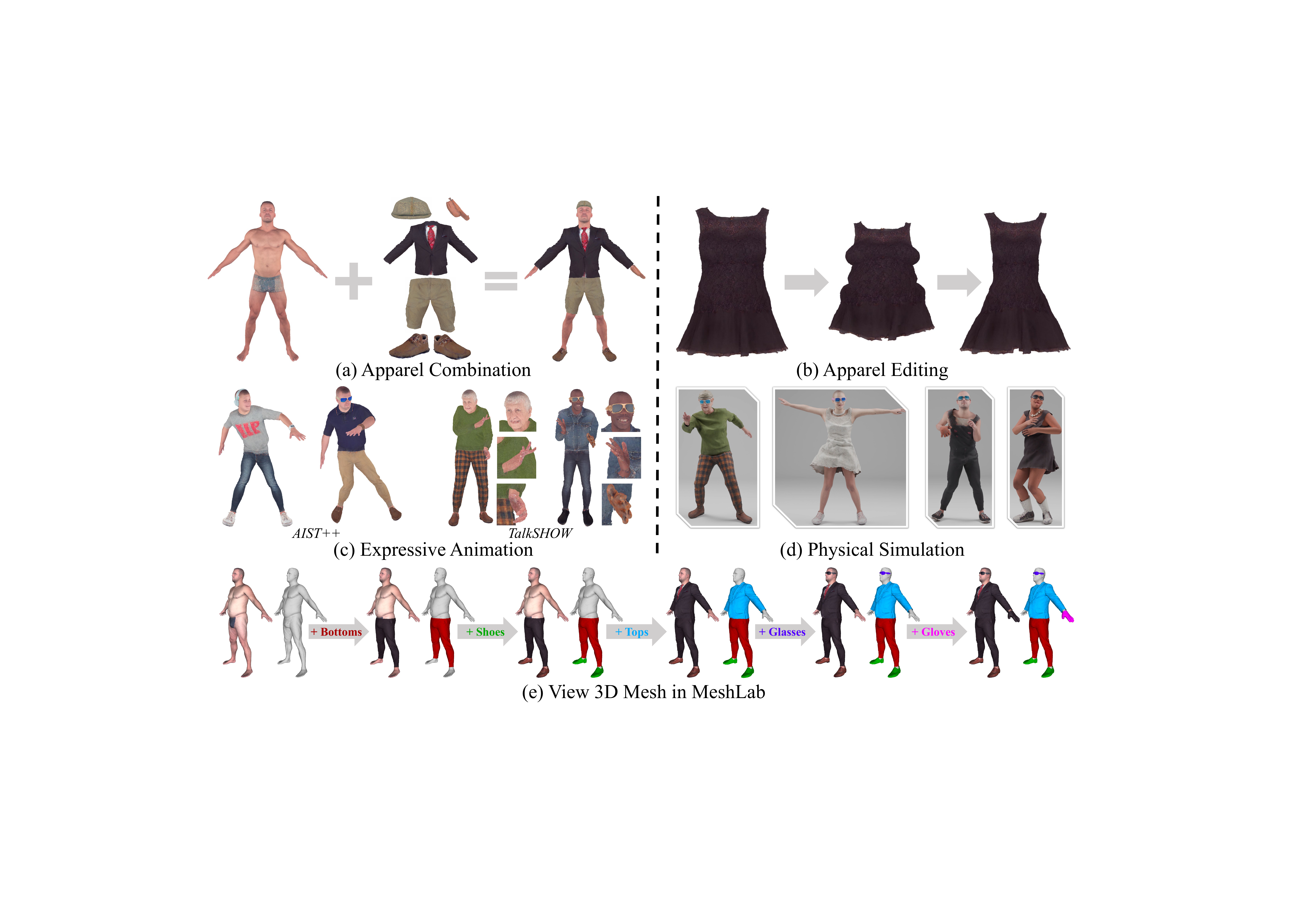}
    \caption{Applications of \blue{DreamBarbie}.}
    \label{fig:app}
    \vspace{-0.5cm}
\end{figure}

\noindent
As shown in the lower part of Table~\ref{tab:quantitative_comparisons}, our approach consistently outperforms all competing methods on all evaluation metrics.
Qualitative comparisons are shown in the right part of Fig.~\ref{fig:comp1}, displaying representative generated outfits.
Compared to other methods, \blue{DreamBarbie} produces apparel with higher-fidelity geometry and textures, while maintaining accurate alignment with the input text, without introducing irrelevant elements such as body parts.
Thanks to the expressive G-Shell~\cite{gshell} representation, we are able to accurately model the non-watertight topology of garments, enabling compatibility with physical simulations (Fig.~\ref{fig:app}-(d)).

\subsection{Ablation Study}
\label{sec:4.4}
\noindent
We conduct detailed ablation studies in Fig.~\ref{fig:ablat} to validate the effectiveness of each component in \blue{DreamBarbie}.

\textit{Effect of SMPLX-Evolving Prior Loss.}
As shown in Fig.~\ref{fig:ablat}-(a), omitting $\mathcal{L}_{prior}$ leads to distorted body proportions and unnatural hands, significantly degrading realism.
Using a fixed human prior ensures anatomical plausibility but results in overly smooth surfaces with limited detail (\textit{e.g.}, missing muscle contours and hair). 
In contrast, the full ${L}_{prior}$ enables both reasonable and detailed human body generation while providing rich semantics that support downstream tasks such as apparel transfer, editing, animation, and simulation.

\textit{Effect of Dual-Space Optimization.}
To generate fine-detailed, well-fitted apparel, we employ dual-space optimization with human- and object-specific generative priors. 
As shown in Fig.~\ref{fig:ablat}-(b), human-space-only optimization yields assets with poor detail (orange box) and erroneously absorbs body parts into clothing (blue box), breaking semantic boundaries. 
In contrast, apparel-space-only optimization produces ill-fitting outfits—evident in oversized shoes (green box) and texture discontinuities near the waist (purple box)—undermining realism. 
Our dual-space approach jointly leverages both priors to achieve detailed, body-conforming garments and visually coherent avatars.

\textit{Effect of Unified Texture Refinement.}
To reduce visual inconsistency between the body and outfits, we introduce a unified texture refinement (UTR) strategy.
Fig.~\ref{fig:ablat}-(c) confirms that UTR effectively reduces texture conflicts across components, thereby enhancing overall realism and aesthetic coherence.

\textit{Effect of Template-Preserving Loss.}
To preserve structural integrity in the generated apparel, we incorporate a template-preserving loss $\mathcal{L}_{{temp}}$.
As shown in Fig.~\ref{fig:ablat}-(d), this loss is essential for preventing unexpected holes and other geometric artifacts that severely degrade apparel quality.

\textit{Effect of Hole-Preserving Loss.}
To accurately model intended garment openings (\textit{e.g.}, necklines), we propose $\mathcal{L}_{{hole}}$.
As demonstrated in Fig.~\ref{fig:ablat}-(e), $\mathcal{L}_{{hole}}$ avoids floating triangles within holes, resulting in clean neckline structures.

\textit{Effect of Collision Loss.}
To avoid intersections between apparel and the underlying body, we introduce $\mathcal{L}_{{colli}}$.
As illustrated in Fig.~\ref{fig:ablat}-(f), neglecting $\mathcal{L}_{{colli}}$ causes severe collision artifacts, which notably compromise result realism.

\subsection{Applications}
\label{sec:4.5}
\noindent
As shown in Fig.~\ref{fig:app}-(a), \blue{DreamBarbie}'s fine-grained decoupling enables the flexible combination of any human body with any outfit, similar to dressing physical Barbie dolls.
Thanks to the SMPLX-evolving prior loss $\mathcal{L}_{prior}$, we achieve accurate alignment between the generated assets and the SMPL-X model. 
This alignment allows for intuitive apparel editing by adjusting shape parameters, and supports natural full-body animation through control of pose and expression parameters, as shown in Fig.~\ref{fig:app}-(b) and -(c). The expression sequences and body motions are taken from AIST++~\cite{aist++} and TalkShow~\cite{talkshow}.
Furthermore, leveraging the expressive G-Shell~\cite{gshell} representation, our method accurately captures non-watertight garment topology, supporting physical simulation~\cite{houdini} as shown in Fig.~\ref{fig:app}-(d).
Our method enables the creation of Barbie-style 3D avatar meshes.
As a result, users can view and combine human bodies and outfits within professional 3D software~\cite{meshlab}, as illustrated in Fig.~\ref{fig:app}-(e).

\begin{figure}[tp]
    \centering
    \includegraphics[width=0.45\textwidth]{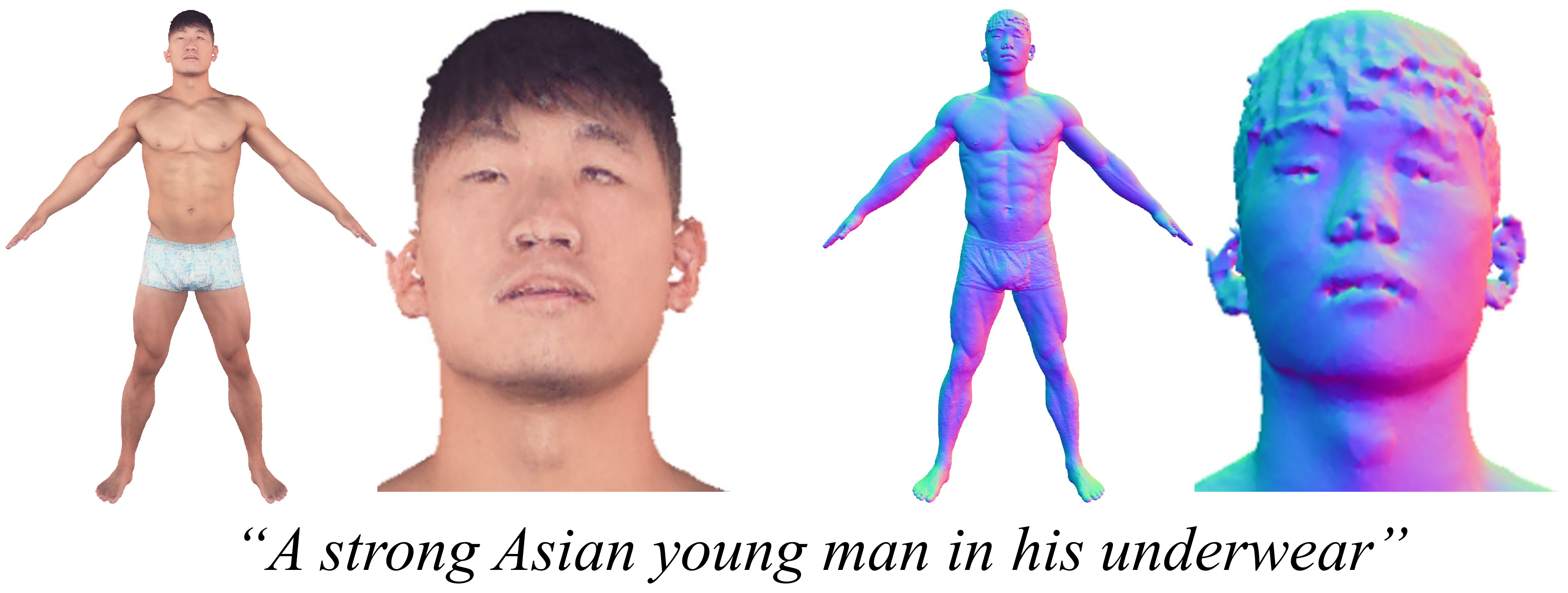}
    \caption{Failure cases.}
    \label{fig:fail}
    \vspace{-0.5cm}
\end{figure}

\vspace{-0.2cm}
\section{limitations and Future Work}
\label{sec:5}
\noindent
Limited by the resolution of G-Shell, our method struggles to accurately model highly complex geometries such as hair, ears, and other facial features, as shown in Fig.~\ref{fig:fail}.
In addition, while the use of predefined templates ensures reasonable initialization, it may slightly limit generation diversity. In future work, we aim to explore template-free, disentangled generation by leveraging 3D generative priors~\cite{trellis3d}.
Finally, our reliance on the SDS loss introduces significant computational overhead, limiting practical applicability. As a direction for future work, we plan to adopt a feed-forward, disentangled generation framework such as StdGen~\cite{stdgen}, integrating diverse human datasets and advanced priors to enable efficient 3D Barbie-style avatar generation.

\vspace{-0.3cm}
\section{Conclusion}
\label{sec:6}
\noindent
We propose \blue{DreamBarbie}, a novel text-guided framework for creating an animatable 3D avatar dressed in decoupled shoes and accessories, along with simulation-ready garments, resembling iconic Barbie dolls.
Specifically, we employ an expressive G-Shell to uniformly represent both watertight and non-watertight meshes.
\blue{We introduce an SDF-based initialization that reformulates boundary definition from manifold geodesics into 3D field intersections, significantly enhancing stability and efficiency of open-surface modeling.}
To guarantee the domain-specific fidelity of the generated human body and outfits, we suitably combine different specific expert T2I models for domain-specific knowledge. 
To mitigate the negative impacts caused by the over-strong generative priors of the expert models, we propose a series of solid regularization losses and optimization strategies to ensure geometric rationality and texture harmony of the generated results. 
Extensive experiments demonstrate that our approach not only outperforms compared methods in both dressed avatar and apparel generation tasks, but also facilitates seamless composition and editing of outfits, as well as expressive animation and physical simulation.

\bibliographystyle{IEEEtran}
\bibliography{paper_ref.bib}


\vspace{-1.5cm}
\begin{IEEEbiography}[{\includegraphics[width=1in,height=1.25in,clip,keepaspectratio]{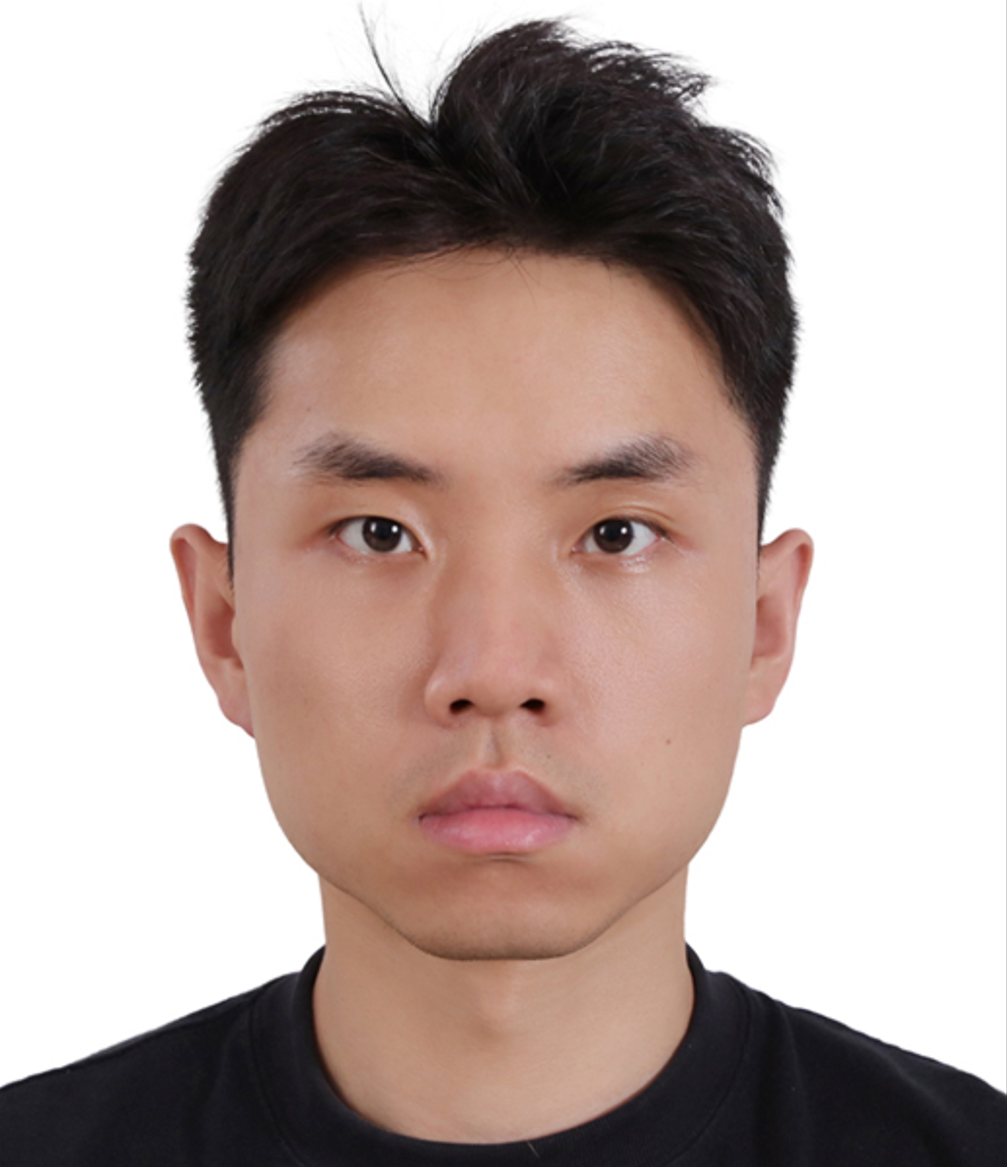}}]{Xiaokun Sun} is currently a Ph.D. candidate at the School of Intelligence Science and Technology, Nanjing University (Suzhou campus). Before this, he received his Master's degree from Tianjin University in 2024. His research interests focus on human-centric 3D digitization.
\end{IEEEbiography}

\vspace{-1.5cm}
\begin{IEEEbiography}[{\includegraphics[width=1in,height=1.25in,clip,keepaspectratio]{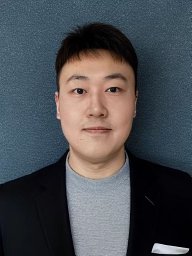}}]{Zhenyu Zhang} is currently an associate professor at the School of Intelligent Science and Technology, Nanjing University (Suzhou campus). He got his Ph.D degree from Nanjing University of Science and Technology in 2020. During 2020-2023, he worked as a staff research scientist at Tencent Youtu Lab. His research interests include 3D modeling, rendering, and generation. His long-term research objective is to create an interactive world simulator from real-world knowledge and human prior.
\end{IEEEbiography}

\vspace{-1.5cm}
\begin{IEEEbiography}[{\includegraphics[width=1in,height=1.25in,clip,keepaspectratio]{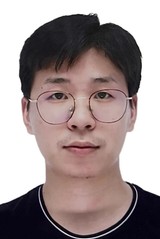}}]{Ying Tai} is currently an associate professor at the School of Intelligence Science and Technology, Nanjing University (Suzhou Campus). He got his Ph.D. degree from Nanjing University of Science and Technology in 2017. During 2017-2023, he worked as a principal researcher and team lead at Tencent Youtu Lab. His research interests include Frontier Generative AI research and applications based on large vision and language models.
\end{IEEEbiography}

\vspace{-1.5cm}
\begin{IEEEbiography}[{\includegraphics[width=1in,height=1.25in,clip,keepaspectratio]{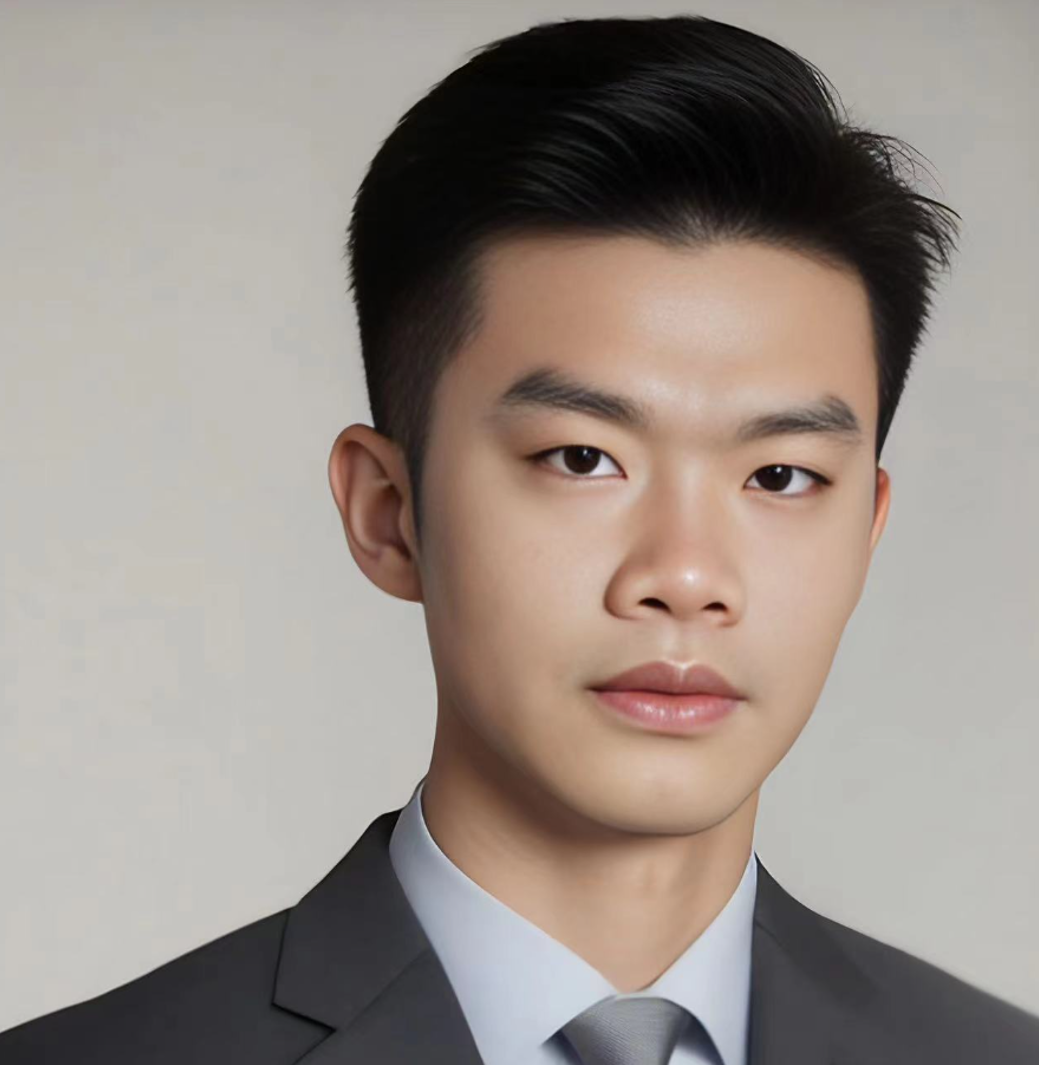}}]{Hao Tang} is a tenure-track Assistant Professor at Peking University, China. He earned a master's degree from Peking University, China, and a Ph.D. from the University of Trento, Italy. His research interests include AIGC, LLM, machine learning, computer vision, embodied AI, and their applications to scientific domains.
\end{IEEEbiography}

\vspace{-1.5cm}
\begin{IEEEbiography}[{\includegraphics[width=1in,height=1.25in,clip,keepaspectratio]{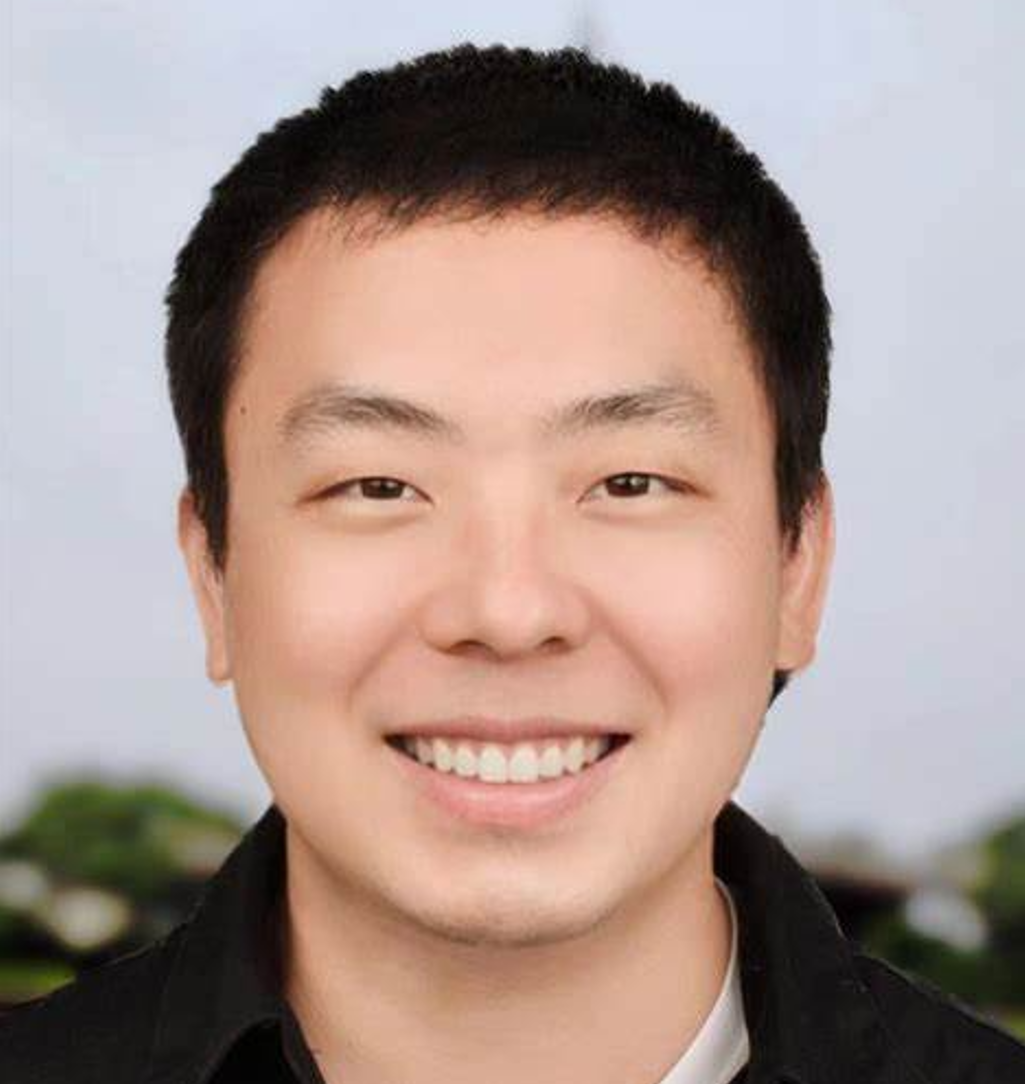}}]{Zili Yi} is currently an associate professor at the School of Intelligence Science and Technology, Nanjing University (Suzhou Campus). He got his Ph.D. degree from Memorial University of Newfoundland, Canada, in 2018. His main research interests include high-quality visual content generation, image/video editing, and multimodal controllable generation.
\end{IEEEbiography}

\vspace{-1.5cm}
\begin{IEEEbiography}[{\includegraphics[width=1in,height=1.25in,clip,keepaspectratio]{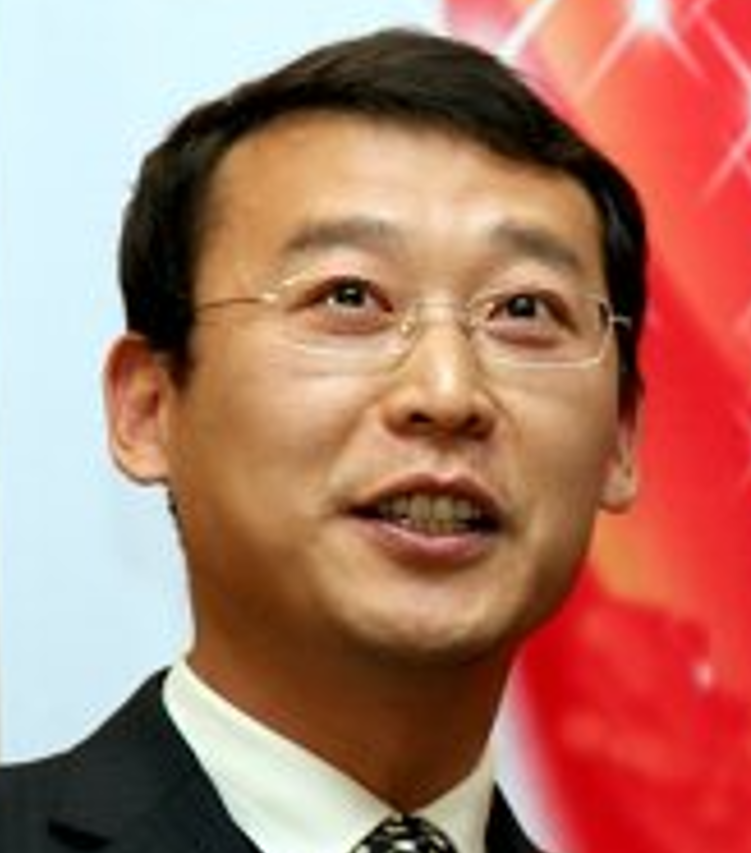}}]{Jian Yang} got his Ph.D. degree from Nanjing University of Science and Technology in 2002. He has authored more than 300 scientific papers in pattern recognition and computer vision. His papers have been cited more than 55000 times in Scholar Google. His research interests include pattern recognition, computer vision, and machine learning.
\end{IEEEbiography}

\section{Supplemental Material Overview}
In this appendix, we consolidate the supplementary content:
\begin{itemize}
    \item Implementation Details.
    \item More Experiment Details and Results.
    \item Prompt List.
    \item Ethics Statement.
\end{itemize}

\section{Implementation Details}
\noindent
\textit{Hyper-Parameters.}
For human body generation, our method requires 6,000 iterations for geometry (the first 300 iterations are used to optimize the shape parameters of the SMPL-X model) and 10,000 iterations for texture. For apparel generation, it requires 5,000 iterations for geometry, 2,000 iterations for texture, and an additional 1,000 iterations to fine-tune the appearance of all garments and accessories, improving overall harmony.
The loss weights $\lambda_{\text{prior}}$, $\lambda_{\text{temp}}$, $\lambda_{\text{hole}}$ and $\lambda_{\text{colli}}$ are set to $1 \times 10^{3}$, $1 \times 10^{2}$, $1 \times 10^{2}$, and $1 \times 10^{4}$, respectively.
For accessories and shoes with simple geometries, we set their G-Shell resolution to $256^{3}$.
For more complex geometries such as human bodies and clothes, we utilize a compact shell-based tetrahedral grid like TeCH~\cite{tech}. 
Further details can be found in the original paper~\cite{tech}.
All experiments are conducted on a single A6000 GPU.
Generating a human body takes roughly 3 hours with 24GB memory, and generating an outfit takes roughly 4 hours with 40GB memory. 

\noindent
\textit{Details of Human Body Generation.}
To improve the stability of the generated human body geometry, we follow the progressive positional encoding of HumanNorm~\cite{humannorm}.
For texture generation, we employ the standard SDS loss during the first 2,000 iterations. In the subsequent 8,000 iterations, we switch to the MSDS loss to further improve the naturalness and realism of the generated appearance.

\noindent
\textit{Details of The SMPLX-Evolving Prior Loss.}
In the SMPLX-evolving prior loss, we periodically optimize $\hat{M}_\text{init}$ to fit the current human body geometry at every 1000 iterations, thus providing an effective and flexible human body prior constraint.
The fitting loss function is as follows:
\begin{equation}
    \mathcal{L}_\text{fit} = \lambda_{\text{chamf}} \mathcal{L}_\text{chamf} + \lambda_{\text{edge}} \mathcal{L}_\text{edge} + \lambda_{\text{nor}} \mathcal{L}_\text{nor} + \lambda_{\text{lap}} \mathcal{L}_\text{lap},
\end{equation}
where $\mathcal{L}_\text{chamf}$ is the Chamfer distance between $\hat{M}_\text{init}$ and the current human body geometry, $\mathcal{L}_\text{edge}$ is the edge length regularization loss, $\mathcal{L}_\text{nor}$ is the normal consistency loss, and $\mathcal{L}_\text{lap}$ is the Laplacian smoothness loss.
The loss weights $\lambda_{\text{chamf}}$, $\lambda_{\text{edge}}$, $\lambda_{\text{nor}}$, and $\lambda_{\text{lap}}$ are set to $1 \times 10^{2}$, $1 \times 10^{0}$, $1 \times 10^{-2}$, and $1 \times 10^{-1}$, respectively.

\noindent
\textit{Details of Fitting The Pie Mesh.}
For each hole boundary, we define a localized, watertight proxy $M_{pie}$ using a parametric elliptical cylinder. 
Specifically, given a point cloud $\mathcal{P}_{hole}$ representing the hole boundary, we perform Principal Component Analysis (PCA) on its coordinates to establish an orthogonal basis $\{ \mathbf{u}, \mathbf{v}, \mathbf{n} \}$. 
The first two components, $\mathbf{u}$ and $\mathbf{v}$, define the optimal fitting plane, while $\mathbf{n}$ represents the surface normal. 
We project $\mathcal{P}_{hole}$ onto this plane to determine the radial extent.
Depending on the boundary topology, we compute the radii $(r_1, r_2)$ by either calculating the maximum radial distance (for circular fitting) or the oriented bounding box dimensions (for elliptical fitting), incorporating a safety margin to ensure full encapsulation. 
The thickness $h$ is derived from the point spread along the normal $\mathbf{n}$.
Finally, $M_{pie}$ is generated by applying a composite affine transformation $\mathbf{T} = \mathbf{T}_{trans} \mathbf{T}_{rot} \mathbf{T}_{scale}$ to a canonical unit cylinder, effectively mapping it to the hole's position, orientation, and scale.

\begin{figure}[tp]
    \centering
    \includegraphics[width=0.45\textwidth]{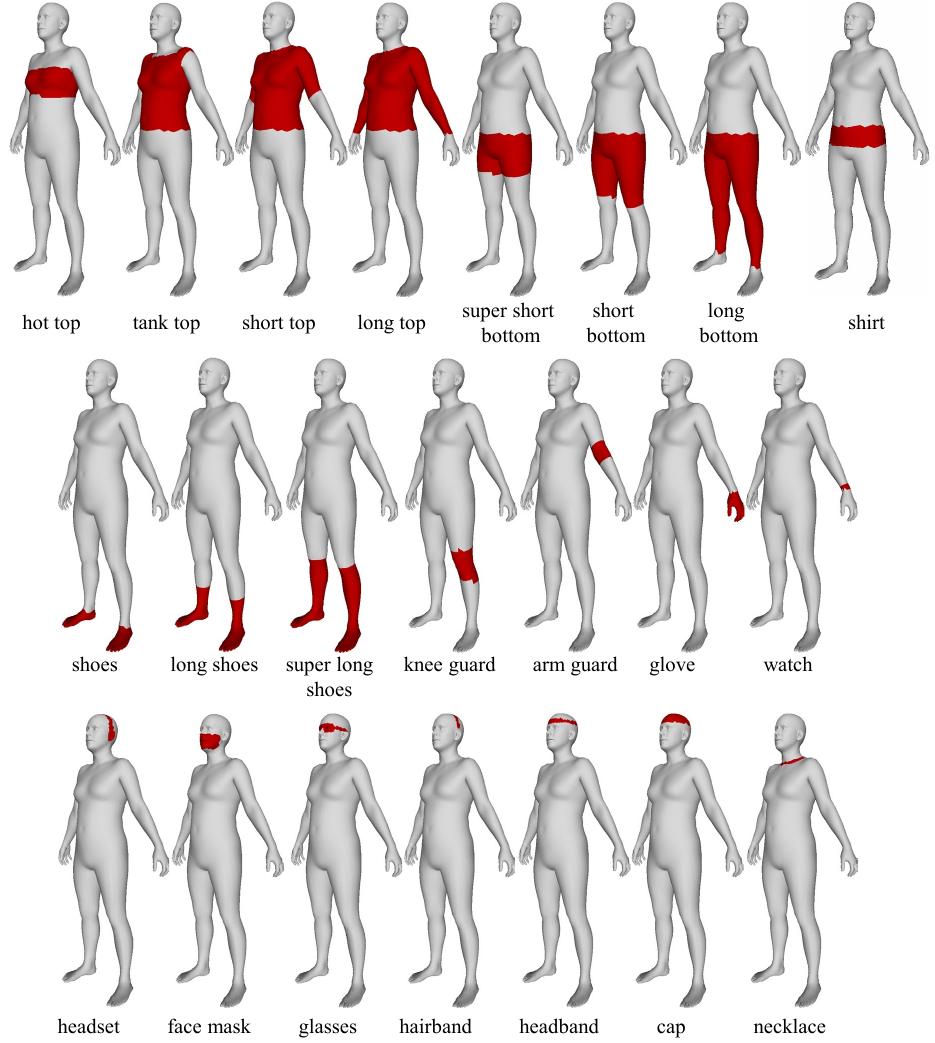} 
    \caption{Illustration of SMPL-X masks for various apparel types.}
    \label{fig:temp}
\end{figure}

\noindent
\textit{Speed Comparison: Geodesic-Based vs. SDF-Based mSDF Initialization.}
(1) Computational Complexity: Calculating the SDF for a spatial point $P$ involves finding the minimum Euclidean distance to the mesh surface. 
By leveraging an R-tree (as implemented in the pysdf library), query complexity is optimized to $O(\log N)$, where $N$ is the number of faces. 
In contrast, calculating the signed geodesic distance requires finding the shortest path constrained to the manifold. 
Even with an efficient Dijkstra-based implementation, the complexity per query reaches $O(V \log V)$, where $V$ is the number of vertices.
(2) Hardware Performance: We evaluated both methods on a single NVIDIA RTX A6000 GPU. 
The initialization process involves 4,000 iterations, with 40,000 points sampled per iteration. 
The SDF-based method completes the task in approximately 3 minutes. 
Conversely, the geodesic-based approach requires nearly 5 hours to achieve the same sampling density. 
Our strategy delivers a speedup of approximately $100\times$, drastically reducing pre-processing time while maintaining superior optimization stability.

\noindent
\textit{SMPL-X Masks.}
As shown in Fig.~\ref{fig:temp}, we predefine SMPL-X masks to cover more than a dozen types of daily clothing and accessories. 
These masks facilitate accurate modeling and alignment with the SMPL-X model.
Additionally, when modeling skirts, we scale the cropped mesh vertically to match the target skirt length.

\section{More Experiment Details and Results}

\begin{figure*}[tp]
    \centering
    \includegraphics[width=0.9\textwidth]{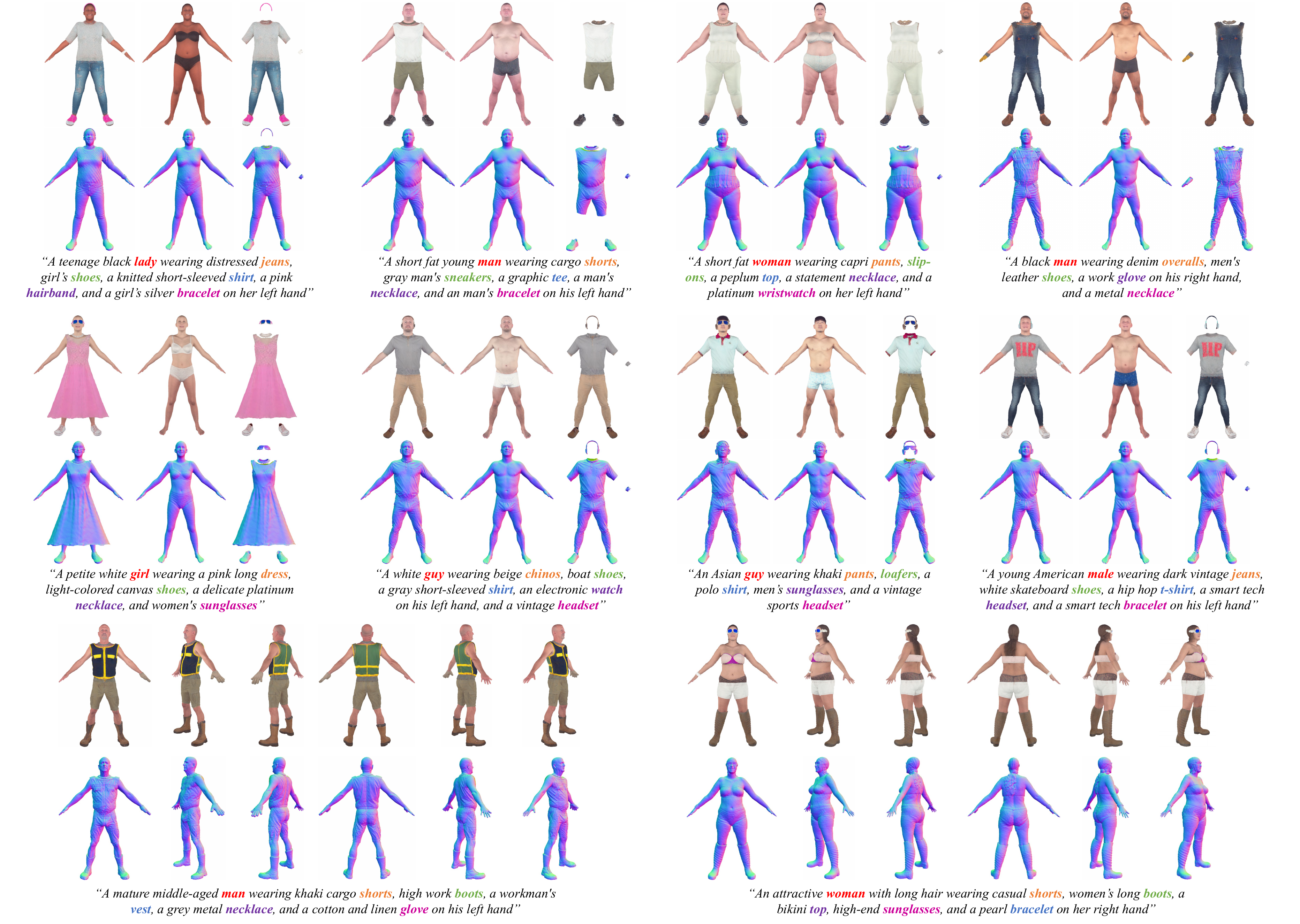} 
    \caption{Qualitative results of our method.}
    \label{fig:show_appendix}
\end{figure*}

\noindent
\textit{Additional Qualitative Results.}
We provide further qualitative results of our method in Fig.~\ref{fig:show_appendix}. These results demonstrate that our approach can generate diverse, high-fidelity bodies and outfits, including tops, bottoms, shoes, glasses, and necklaces.

\begin{figure*}[tp]
    \centering
    \includegraphics[width=0.9\textwidth]{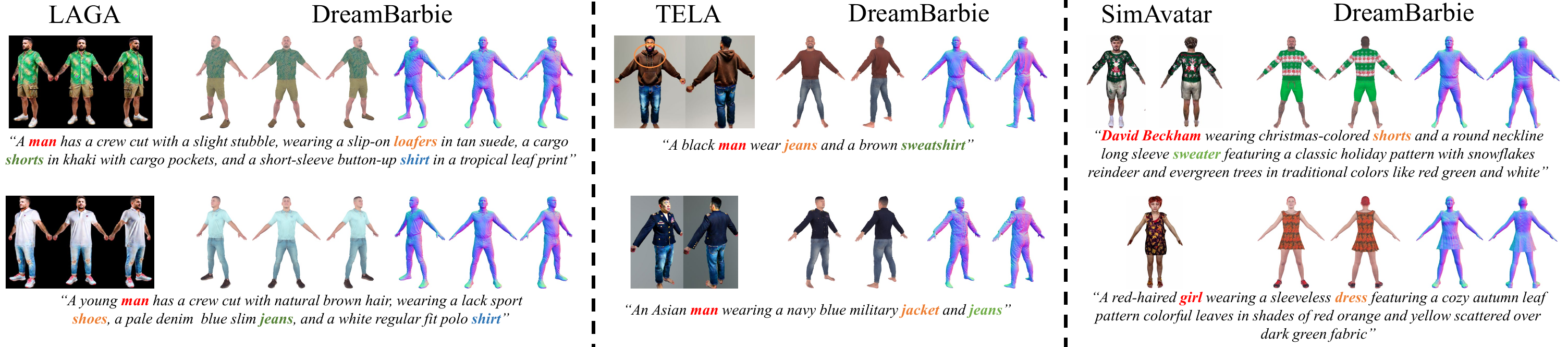} 
    \caption{Qualitative comparisons with LAGA~\cite{laga}, TELA~\cite{tela} and SimAvatar~\cite{simavatar}. The results of LAGA, TELA, and SimAvatar are copied from their papers.}
    \label{fig:comp1_appendix}
\end{figure*}

\noindent
\textit{Comparisons with Text-to-Disentangled-Avatar Methods.}
We compare our method qualitatively with other closed-source text-to-disentangled-avatar approaches: LAGA~\cite{laga}, TELA~\cite{tela}, and SimAvatar~\cite{simavatar}. 
As shown in Fig.~\ref{fig:comp1_appendix}, our method achieves superior performance in generating refined geometry and realistic textures. 
In addition, DreamBarbie supports a wide variety of apparel items, such as tops, bottoms, dresses, shoes, necklaces, and glasses.

\begin{figure}[tp]
    \centering
    \includegraphics[width=0.45\textwidth]{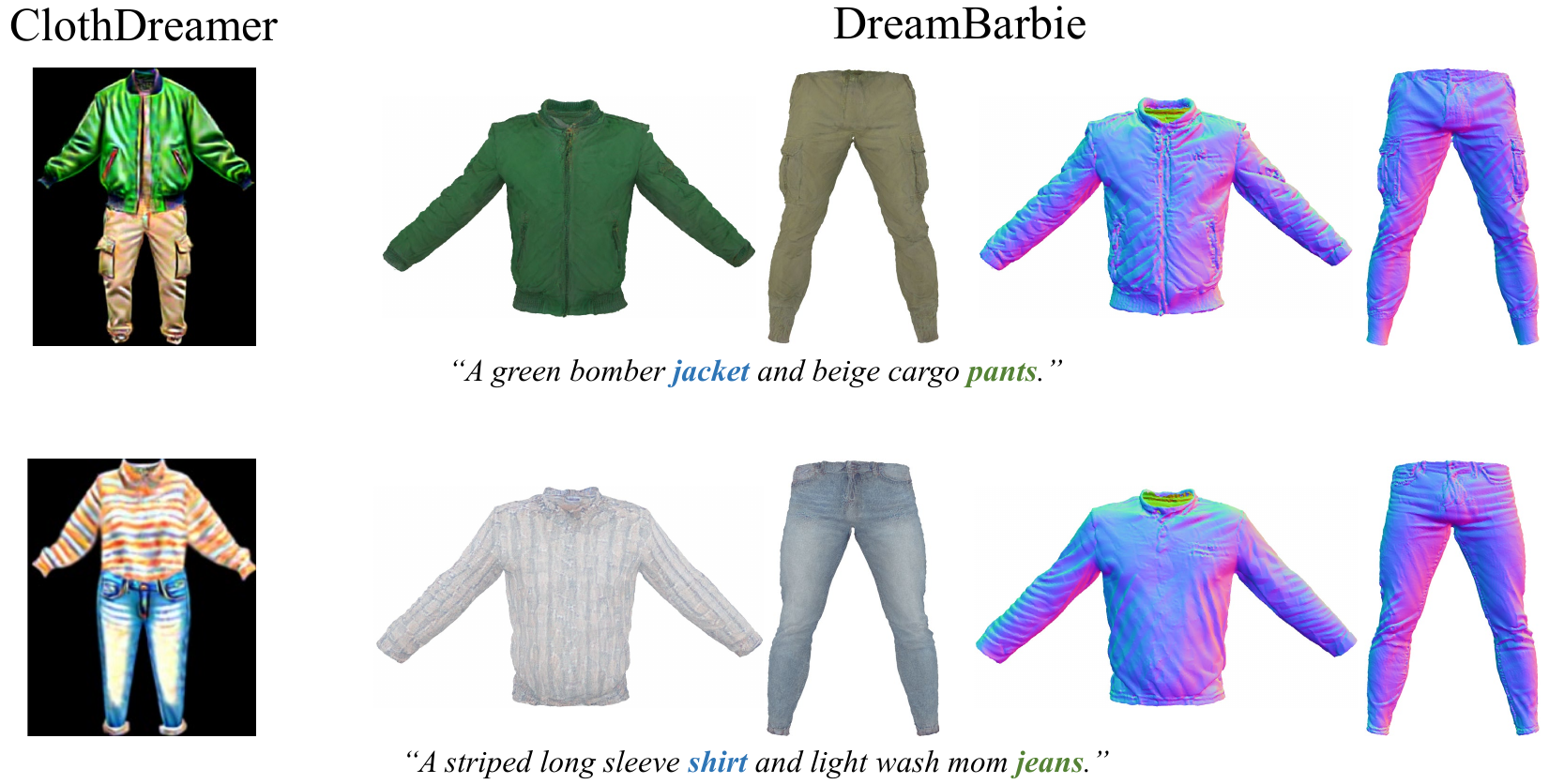} 
    \caption{Qualitative comparisons with ClothDreamer~\cite{clothedreamer}. The results of ClothDreamer are copied from its paper.}
    \label{fig:comp2_appendix}
\end{figure}

\noindent
\textit{Comparisons with Text-to-Garment Methods.}
We present qualitative comparisons with the closed-source text-to-garment work ClothDreamer in Fig.~\ref{fig:comp2_appendix}. 
Compared to ClothDreamer~\cite{clothedreamer}, our method generates more complex geometric details and supports the generation of full-body avatars, including shoes and accessories, offering users an immersive, Barbie-style digital experience.

\noindent
\textit{Disadvantages of CLIP-Based Metrics.}
Recent works~\cite{t2i-compbench,llmscore} have shown that CLIP-based metrics~\cite{clip,openclip,fashionclip} can only measure coarse text-image similarity and fail to accurately measure the fine-grained correspondence between 3D content and input prompts. Consequently, we follow Progressive3D~\cite{progressive3d} and adopt fine-grained text-to-image evaluation metrics, including BLIP-VQA~\cite{blip,lavis} and BLIP2-VQA~\cite{blip2,lavis}, to evaluate the generation capacity of methods.

\begin{figure}[tp]
    \centering
    \includegraphics[width=0.45\textwidth]{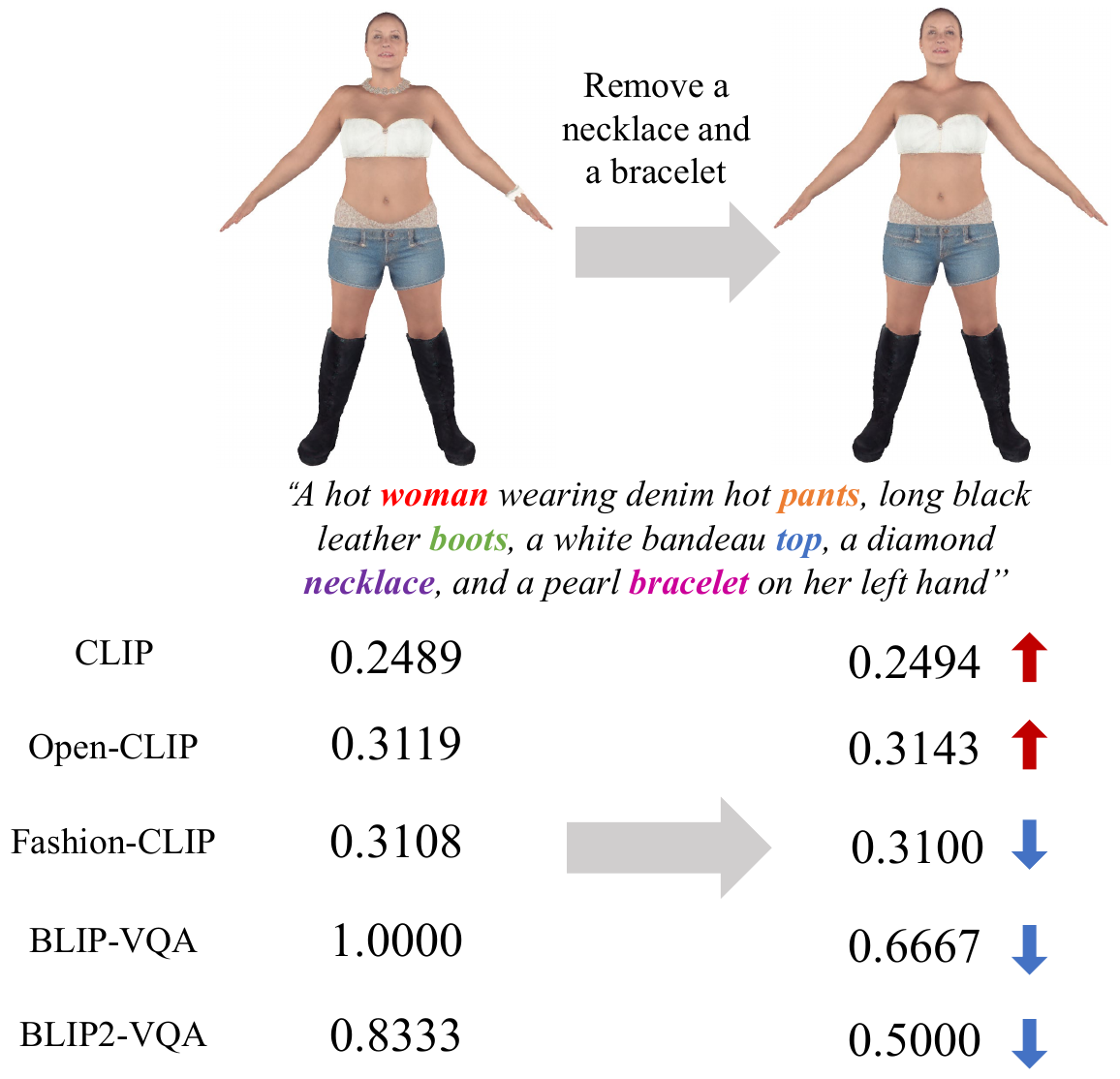} 
    \caption{Quantitative comparisons for metrics including CLIP, Open-CLIP, Fashion-CLIP, BLIP-VQA, and BLIP2-VQA.}
    \label{fig:clip}
\end{figure}

As shown in Fig.~\ref{fig:clip}, under the condition that the input text remains unchanged, the removal of the avatar's necklace and bracelet does not cause significant changes in the CLIP-based metrics; in fact, the scores of CLIP and Open-CLIP even increase. 
However, the scores of BLIP-VQA and BLIP2-VQA decrease significantly, clearly demonstrating that CLIP-based metrics are incapable of measuring fine-grained correspondences, whereas BLIP-VQA and BLIP2-VQA perform well in this aspect.

\begin{table}[H]
    \begin{center}
    \caption{Quantitative performance across different garment types}
    \label{tab:diff_garment}
    \resizebox{0.45\textwidth}{!}{
    \begin{tabular}{c|c|c|c}
    \hline
    Metric & Tight-Fitting Tops & Tight-Fitting Bottoms & Loose Garments \\
    \hline
    BLIP-VQA $\uparrow$ & 0.9750 & 0.9231 & 0.7857 \\
    BLIP2-VQA $\uparrow$ & 0.9250 & 0.8846 & 0.7143 \\
    \hline
    \end{tabular}
    }
    \end{center}
    \vspace{-0.5cm}
\end{table}

\begin{figure}[H]
    \centering
    \includegraphics[width=0.95\linewidth]{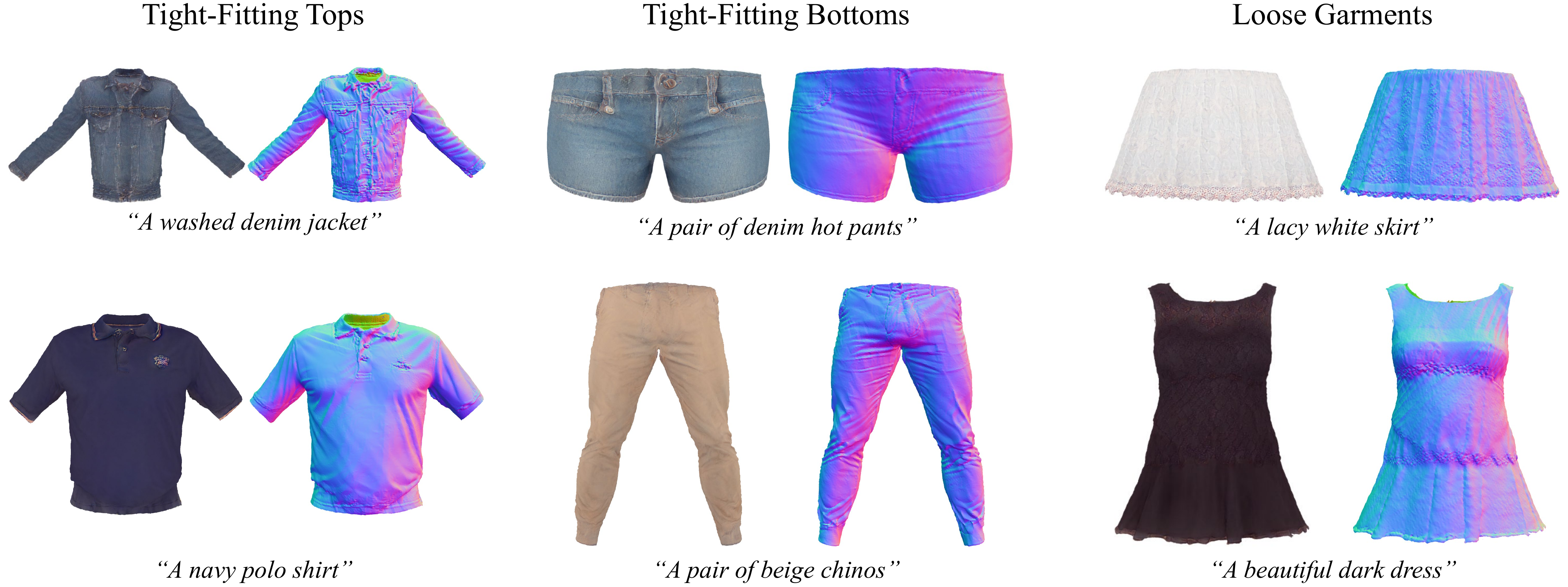}
    \caption{Qualitative comparison of generation results across different garment types: tight-fitting tops, tight-fitting bottoms, and loose garments.}
    \label{fig:diff_garment}
\end{figure}

\noindent
\textit{The Performance Differences across Garment Types.}
We include a quantitative comparison of performance across different garment categories.
Because the original test set contained very few examples of loose clothing, we augmented it with 10 additional samples featuring loose garments (e.g., dresses, skirts) to reduce randomness and improve the reliability of our analysis.
As shown in Table~\ref{tab:diff_garment}, we report BLIP-VQA and BLIP2-VQA scores (see Section IV-A for evaluation details) for three garment categories: tight-fitting tops (e.g., jackets), tight-fitting bottoms (e.g., jeans), and loose garments (e.g., dresses). Figure~\ref{fig:diff_garment} further provides qualitative visual comparisons of generation results across these categories.

The results indicate that the generation quality for loose garments is somewhat lower than that for fitted clothing, as reflected in both reduced VQA scores and fewer fine-grained fabric fold details. 
This challenge is well recognized in the field of disentangled digital human generation. 
Notably, while prior methods such as SO-SMPL~\cite{so-smpl}, TELA~\cite{tela}, and LAGA~\cite{laga} often fail entirely on skirts, our approach demonstrates clear and meaningful progress.
In future work, we plan to incorporate template-free 3D generative priors~\cite{trellis3d} to further enhance the realism and structural fidelity of loose-fitting garment modeling.

\begin{figure}[H]
    \centering
    \includegraphics[width=0.95\linewidth]{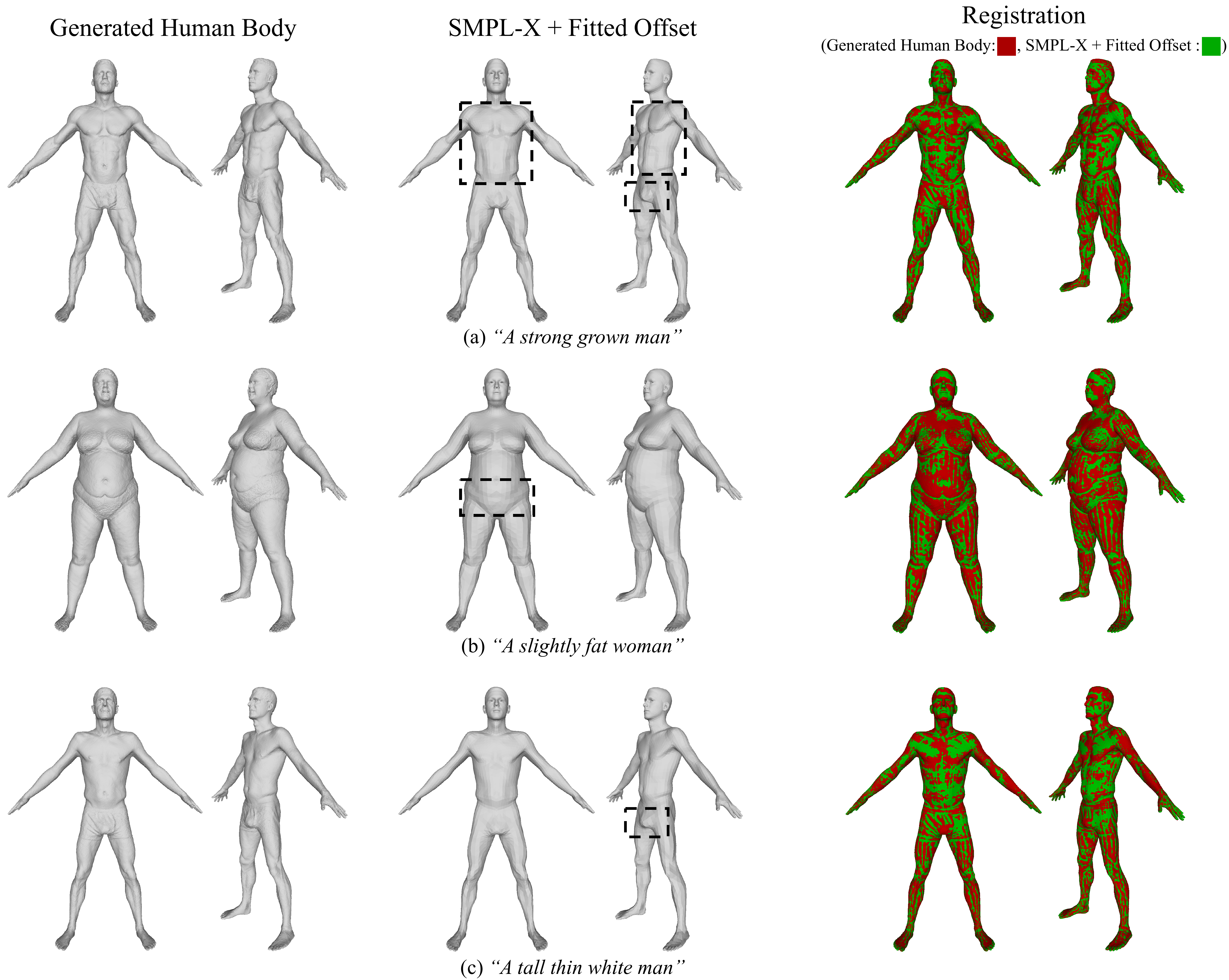}
    \caption{Robust SMPL-X fitting results across various genders and body types. The box highlights detailed geometric features such as muscle lines, waist fat, and male genitalia.}
    \label{fig:fitting}
\end{figure}

\noindent
\textit{Robustness of SMPL-X Fitting in the SMPLX-Evolving Prior Loss.}
Unlike conventional SMPL-X registration—which jointly optimizes shape, pose, global rotation, translation, scale, and per-vertex offsets—SMPL-X Fitting in our SMPLX-evolving prior loss only optimizes the per-vertex offsets. 
This significantly simplifies the optimization problem.
Moreover, because $\theta_h$ is initialized from $M_{\text{init}}$, it remains closely aligned with $\hat{M}_{\text{init}}$ at the start of optimization. 
The evolving prior loss further enforces structural plausibility, greatly reducing the likelihood of registration failure.
As demonstrated in Figure~\ref{fig:fitting}, our method achieves highly accurate SMPL-X fits across various genders and body types, with an average Chamfer distance of merely $0.000107$ on the test set.
Fine geometric details—including muscle definition, waist adiposity, and male genitalia—are faithfully reconstructed, as highlighted in the box in Figure~\ref{fig:fitting}.
In summary, our SMPL-X registration is exceptionally robust, obviating the need for a fallback strategy.

\section{Prompt List}
The following are textual prompts for quantitative experiments:
\begin{itemize}
    \item A strong grown man wearing cargo pants, brown leather boots, a red plaid flannel shirt, a work glove on his left hand, and a silver wristwatch on his right hand.
    \item A thin boy wearing blue denim shorts, white high-top sneakers, a cotton t-shirt, a navy baseball cap, and a sports watch on his left hand.
    \item A normal-sized American man wearing beige khaki pants, suede loafers, a navy polo shirt, aviator sunglasses, and a leather wristband on his left hand.
    \item A black African man wearing mesh basketball shorts, high-top basketball shoes, a basketball vest, a sports knee guard on his left knee, and a sports arm guard on his right arm.
    \item A little yellow girl wearing pink floral leggings, white ballet flats, a cotton short sleeve, myopia glasses, and a charm bracelet on her left hand.
    \item A fat European man wearing grey sweatpants, black running shoes, a cotton hoodie, a knit beanie, and a black mask.
    \item A tall, thin African man wearing dark blue jeans, black cow shoes, a washed denim jacket, a leather glove on his left hand, and gold-rimmed sunglasses.
    \item A white European woman wearing black tailored trousers, nude patent heels, a white blouse, a mask, and women's sunglasses.
    \item A strong woman wearing black jogger pants, grey trainers, a red tank top, a sports headset, and a sports necklace.
    \item A skinny boy wearing army green overalls, white canvas shoes, a baseball jacket, a baseball cap, and a baseball glove on his left hand.
    \item A white guy wearing beige chinos, boat shoes, a gray short-sleeved shirt, an electronic watch on his left hand, and a vintage headset.
    \item A mature, middle-aged man wearing khaki cargo shorts, high work boots, a workman's vest, a grey metal necklace, and a cotton and linen glove on his left hand.
    \item A young American male wearing dark vintage jeans, white skateboard shoes, a hip hop t-shirt, a smart tech headset, and a smart tech bracelet on his left hand.
    \item A hot woman wearing denim hot pants, long black leather boots, a white bandeau top, a diamond necklace, and a pearl bracelet on her left hand.
    \item An underage boy wearing grey sweat shorts, black trainers, a blue tee, a student watch on his left hand, and a beret.
    \item An old European grandmother wearing brown plaid pants, leather loafers, a green wool sweater, a flat cap, and reading glasses.
    \item A black boy wearing black track pants, white sneakers, a red hoodie, a medical mask, and a sports glove on his left hand.
    \item A black woman wearing snow pants, snow boots, a snow down jacket, a snow glove on her left hand, and snow goggles.
    \item An old Asian grandfather wearing linen trousers, flat shoes, a cotton shirt, a jade bracelet on his left hand, and a gold bracelet on his right hand.
    \item A big American man wearing black suit pants, leather shoes, a suit with a tie, dark sunglasses, and a black leather glove on his left hand.
    \item A strong Asian young man wearing quick-drying gym shorts, gray sneakers, a quick-drying gym vest, a gold necklace, and an exercise bracelet on his left hand.
    \item A petite white girl wearing a short pink floral skirt, ballet flats, a lace dance vest, a pearl necklace, and a platinum bracelet on her left hand.
    \item A normal-sized black man wearing nylon gym shorts, badminton shoes, a nylon gym vest, a cotton sports headband, and a black sports arm guard on his left arm.
    \item An African woman wearing a short black sports skirt, vintage running shoes, a sports bandeau top, a red sports headband, and a cotton sports knee guard on her left knee.
    \item A slim yellow woman wearing silk pants, red high heels, a silk blouse, black-rimmed glasses, and a surgical mask.
    \item A little boy wearing blue beach shorts, cotton white socks, a white t-shirt, a child's watch on his left hand, and children's glasses.
    \item A teenage girl wearing a long lacy white skirt, dance shoes, a knitted short-sleeved shirt, a pink hairband, and a silver bracelet on her left hand.
    \item A teenage boy wearing gray basketball shorts, high-top black socks, a white beach vest, silver goggles, and a rubber sports bracelet on his left hand.
    \item A middle-aged woman wearing a long silk skirt, beige high heels, a black silk blouse, a white floral hairband, and an emerald bracelet on her left hand.
    \item A slightly fat man wearing straight-leg jeans, white air force one sneakers, a brown cotton coat, a gray beanie hat, and vintage sunglasses.
\end{itemize}

\section{Ethics Statement}
DreamBarbie offers users a low-cost approach to creating high-quality, disentangled avatars with fine-grained control over various attributes such as body shape, clothing, and accessories. 
This makes avatar generation more accessible to non-experts and reduces the need for specialized tools or manual modeling. 
However, the ability to generate realistic and customizable 3D avatars from text also introduces potential risks, such as the misuse of generated content for deceptive or harmful purposes. 
Similar concerns are commonly seen in other AI-generated media, including deepfakes and synthetic images. 
As generative models continue to advance, it is crucial for researchers and developers to proactively investigate these ethical challenges and develop strategies to mitigate associated risks.

\end{document}